\PassOptionsToPackage{pdftex}{graphicx}
\documentclass[10pt,twocolumn,letterpaper]{article}

\usepackage[pagenumbers]{iccv} % To force page numbers, e.g. for an arXiv version

\definecolor{iccvblue}{rgb}{0.21,0.49,0.74}
\usepackage{bm}
\usepackage{multirow}

\usepackage{amsmath}
\usepackage{amssymb}

\usepackage{xcolor}
\usepackage[pdftex]{graphicx}

\usepackage[pagebackref,breaklinks,colorlinks,allcolors=iccvblue]{hyperref}
\title{Semi-supervised Concept Bottleneck Models}

\author{
\textbf{Lijie Hu$^{1,2}$, Tianhao Huang$^{1,2,3}$, Huanyi Xie$^{1,2,4}$, Xilin Gong$^{1,2,5}$}\\
\textbf{Chenyang Ren$^{1,2,6}$, Zhengyu Hu$^{1,2,7}$, Lu Yu$^{8}$, Ping Ma$^{5}$, and Di Wang$^{1,2}$}\\
$^1$Provable Responsible AI and Data Analytics (PRADA) Lab\\
$^2$KAUST \quad $^3$University of Virginia \quad  $^4$KTH Royal Institute of Technology  \\ 
 $^5$ University of Georgia \quad $^6$Shanghai Jiao Tong University \quad $^7$HKUST \quad $^8$Ant Group
}

\begin{document}
\maketitle
\begin{abstract} 
Concept Bottleneck Models (CBMs) have garnered increasing attention due to their ability to provide concept-based explanations for black-box deep learning models while achieving high final prediction accuracy using human-like concepts. However, the training of current CBMs is heavily dependent on the precision and richness of the annotated concepts in the dataset. These concept labels are typically provided by experts, which can be costly and require significant resources and effort. Additionally, concept saliency maps frequently misalign with input saliency maps, causing concept predictions to correspond to irrelevant input features - an issue related to annotation alignment. To address these limitations, we propose a new framework called \textbf{SSCBM} (\underline{\textbf{S}}emi-\underline{\textbf{s}}upervised \underline{\textbf{C}}oncept \underline{\textbf{B}}ottleneck \underline{\textbf{M}}odel). Our SSCBM is suitable for practical situations where annotated data is scarce. By leveraging joint training on both labeled and unlabeled data and aligning the unlabeled data at the concept level, we effectively solve these issues. We proposed a strategy to generate pseudo labels and an alignment loss. Experiments demonstrate that our SSCBM is both effective and efficient. With only 10\% labeled data, our model's concept and task accuracy on average across four datasets is only 2.44\% and 3.93\% lower, respectively, compared to the best baseline in the fully supervised learning setting.
\end{abstract}
\section{Introduction}
Recently, deep learning models, such as ResNet \cite{he2016deep}, often feature complex non-linear architectures, making it difficult for end-users to understand and trust their decisions. This lack of interpretability is a significant obstacle to their adoption, especially in critical fields such as healthcare \cite{thirunavukarasu2023large} and finance \cite{li2023large}, where transparency is crucial. Explainable artificial intelligence (XAI) models have been developed to meet the demand for transparency, providing insights into their behavior and internal mechanisms \cite{hu2023seat,hu2023improving,yang2024human,hu2024hopfieldian}. Concept Bottleneck Models (CBMs) \cite{koh2020concept} are particularly notable among these XAI models for their ability to clarify the prediction process of end-to-end AI models. CBMs introduce a bottleneck layer that incorporates human-understandable concepts. During prediction, CBMs first predict concept labels from the original input and then use these predicted concepts in the bottleneck layer to determine the final classification label. This approach results in a self-explanatory decision-making process that users can comprehend.

\begin{figure*}[ht]
\centering
\resizebox{0.9\linewidth}{!}{
\begin{tabular}{ccc}
\includegraphics[bb=0 0 600 600, width=0.33\textwidth]{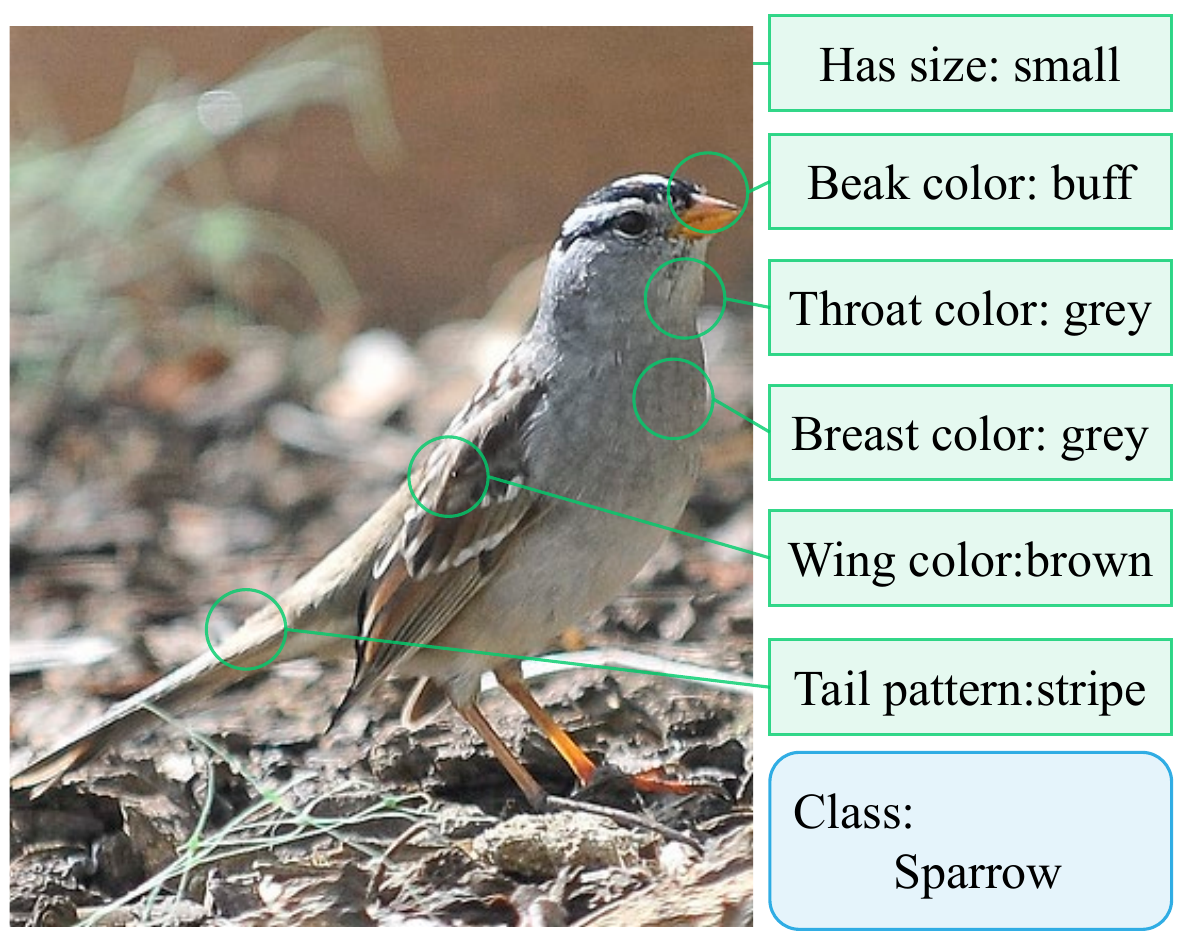} & 
\includegraphics[bb=0 0 700 700, width=0.33\textwidth]{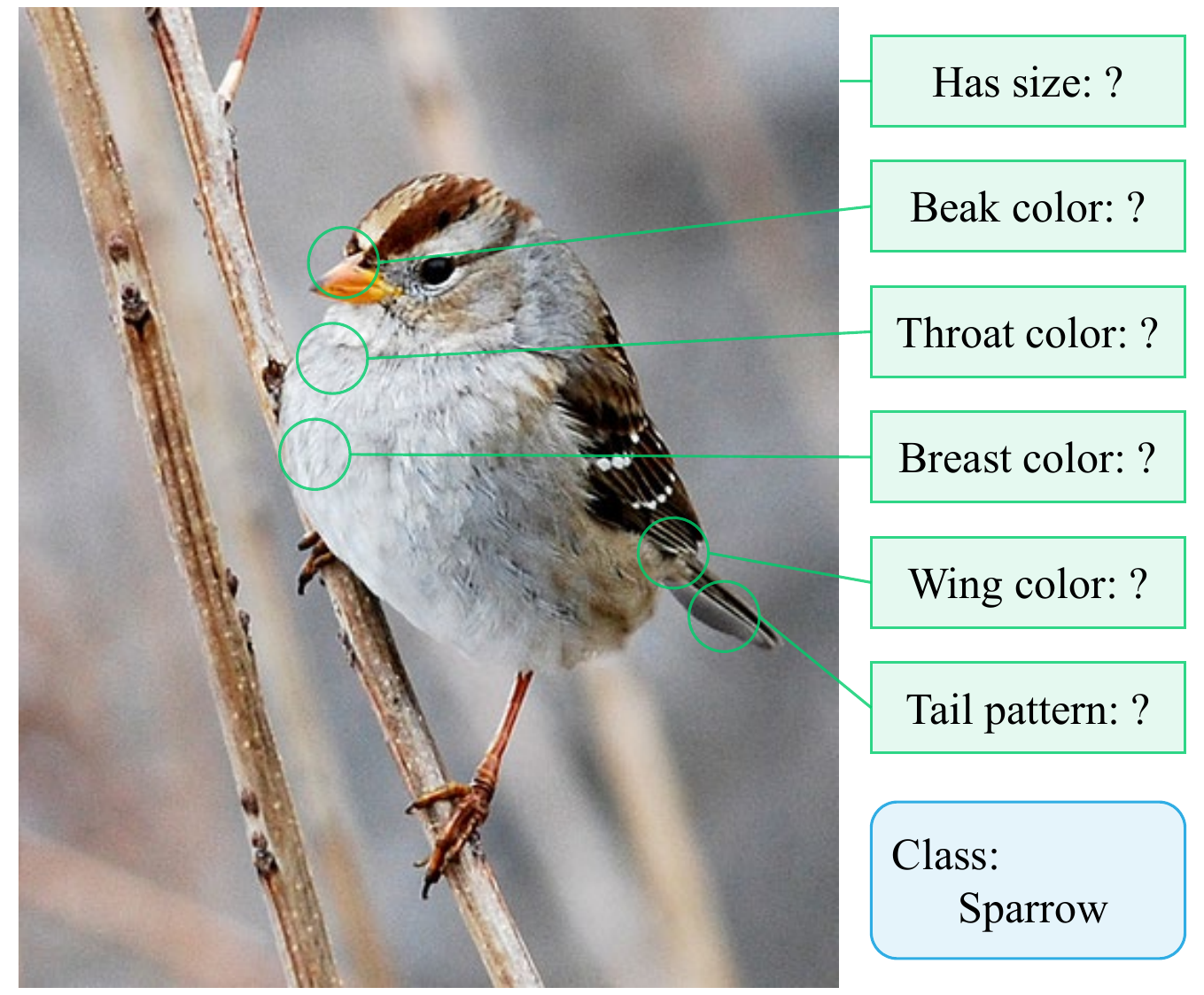} & 
\includegraphics[width=0.33\textwidth]{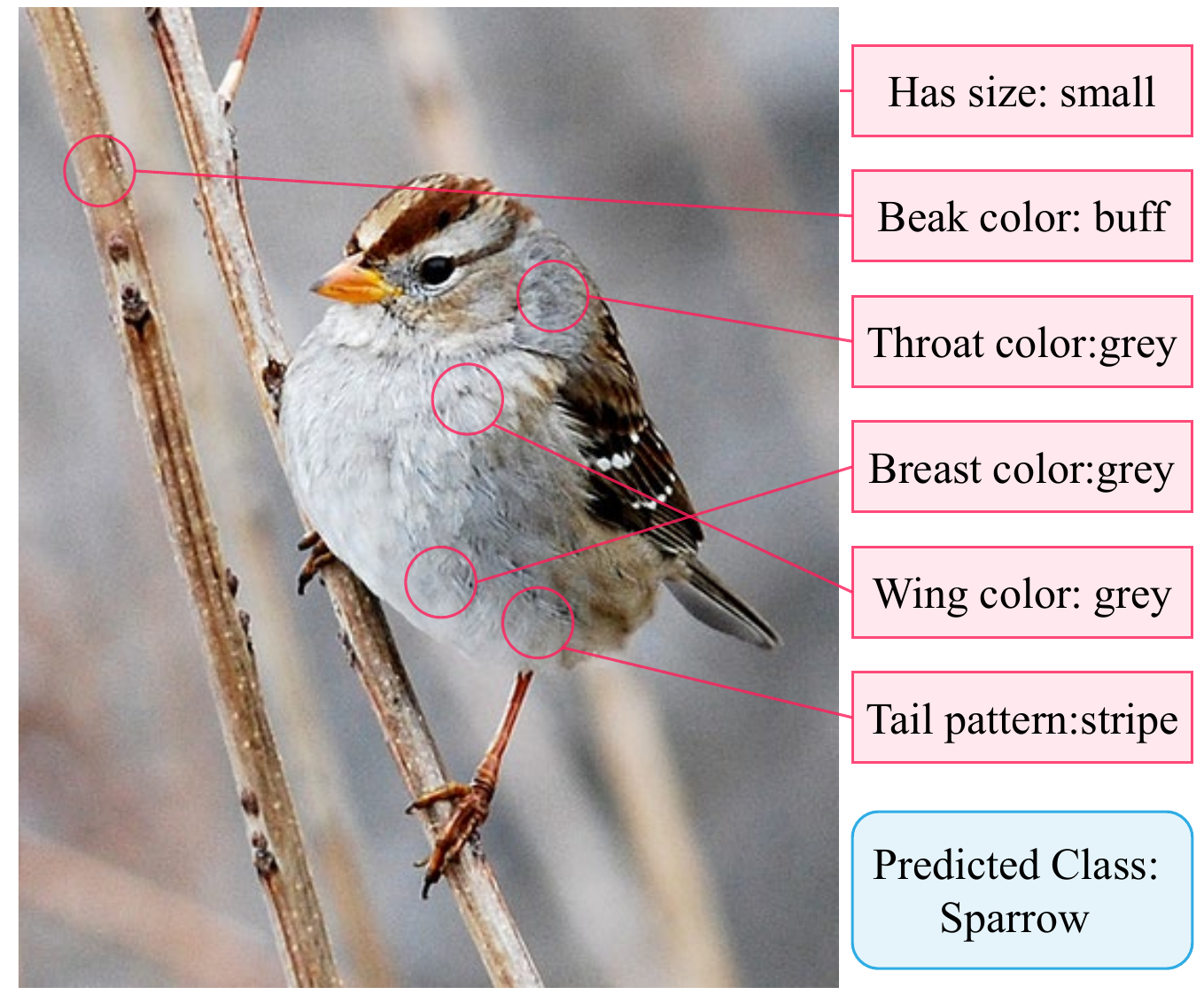} \\
(a) Label Complete and Well-aligned  & (b) Label Incomplete & (c) Misaligned
\end{tabular}}
\caption{(a) A sample of sparrow class with complete concept labels. (b) A sample of sparrow class with incomplete concept labels. (c) A sample of misalignment between input features and concepts resulting from existing CBM methods. Our framework simultaneously utilizes both (a) and (b) types of data and addresses the issue of (c) through an alignment loss. \label{fig:intro}}
\end{figure*}

A major issue in original CBMs is the need for expert labeling, which is costly in practice. Some researchers address this problem through unsupervised learning. For example, \cite{oikarinen2023label} proposes a Label-free CBM that transforms any neural network into an interpretable CBM without requiring labeled concept data while maintaining high accuracy. Similarly, Post-hoc Concept Bottleneck models \cite{yuksekgonul2022post} can be applied to various neural networks without compromising performance, preserving interpretability advantages. However, these methods have three issues. First, those unsupervised methods heavily rely on large language models like GPT-3, which have reliability issues \cite{lai2023faithful}. Second, the concepts extracted by these models lack evaluation metrics, undermining their interpretability. Third, the assumption that no concept labels are available is too stringent in practice. In reality, obtaining a small portion of concept labels is feasible and cost-effective. Therefore, we can maximize the use of this small labeled concept dataset. This is the motivation for introducing our framework, which focuses on the semi-supervised setting in CBM.

In this paper, we introduce a framework called the \textbf{SSCBM} (\underline{\textbf{S}}emi-\underline{\textbf{s}}upervised \underline{\textbf{C}}oncept \underline{\textbf{B}}ottleneck \underline{\textbf{M}}odel). Compared to a supervised setting, semi-supervised  CBMs have two main challenges. First, obtaining concept embeddings requires concept labels, so we need to generate pseudo labels to obtain these concept embeddings. To achieve this, SSCBM uses a KNN-based algorithm to assign pseudo-concept labels for unlabeled data.

Second, while such a simple pseudo-labeling method is effective and has acceptable classification accuracy, we also find that the concept saliency map often misaligns with the input saliency map, meaning concept predictions frequently correspond to irrelevant input features.  This misalignment often arises from inaccurate concept annotations or unclear relationships between input features and concepts, which is closely related to the broader issue of annotation alignment. In fact, in the supervised setting, there is a similar misalignment issue \cite{furby2024can}. Existing research seeks to improve alignment by connecting textual and image information \cite{hou2024concept}. However, these methods only focus on supervised settings and cannot be directly applied to our settings, as our pseudo-labels contain noise. Our framework achieves excellent performance in both concept accuracy and concept saliency alignment by leveraging joint training on both labeled and unlabeled data and aligning the unlabeled data at the concept level. To achieve this, we leverage the relevance between the input image and the concept and get other pseudo-concept labels based on these similarity scores. Finally, we align these two types of pseudo-concept labels to give the concept encoder the ability to extract useful information from features while also inheriting the ability to align concept embeddings with the input.  See Figure \ref{fig:intro} for an illustration. 

Comprehensive experiments on benchmark datasets demonstrate that our SSCBM is efficient and effective. Our contributions are summarized as follows.

\begin{itemize}
    \item We propose SSCBM, a framework designed to solve the semi-supervised annotation problem, which holds practical significance in real-world applications. To the best of our knowledge, we are the first to tackle these two problems within a single framework, elucidating the behavior of CBMs through semi-supervised alignment. 
    \item Our framework addresses the semi-supervised annotation problem alongside the concept semantics alignment problem in a simple and clever manner. We first use the KNN algorithm to assign a pseudo-label to each unlabeled data, which has been
experimentally proven to be simple and effective. Then, we compute a heatmap between 
concept embeddings and the input. After applying a threshold, we obtain the predicted alignment
label. Finally, we optimize the alignment loss between these two pseudo-concept labels to mitigate the misalignment issue. 
    \item Comprehensive experiments demonstrate the superiority of our SSCBM in annotation and concept-saliency alignment, indicating its efficiency and effectiveness. Using only 10\% labeled data, our model achieves concept and task accuracy on average across the four datasets that are merely 2.44\% and 3.93\% lower, respectively, than the best baseline under fully supervised learning settings.
\end{itemize}
\section{Related Work}
\noindent {\bf Concept Bottleneck Models.}
Concept Bottleneck Model (CBM) \cite{koh2020concept} is an innovative deep-learning approach for image classification and visual reasoning. By introducing a concept bottleneck layer into deep neural networks, CBMs enhance model generalization and interpretability by learning specific concepts. However, CBMs face two primary challenges: their performance often lags behind that of models without the bottleneck layer due to incomplete information extraction, and they rely heavily on laborious dataset annotation. Researchers have explored various solutions to these challenges. \cite{chauhan2023interactive} extends CBMs into interactive prediction settings by introducing an interaction policy to determine which concepts to label, thereby improving final predictions. \cite{oikarinen2023label} addresses CBM limitations by proposing a Label-free CBM, which transforms any neural network into an interpretable CBM without requiring labeled concept data, maintaining high accuracy. Post-hoc Concept Bottleneck models \cite{yuksekgonul2022post} can be applied to various neural networks without compromising performance, preserving interpretability advantages. Related work in the image domain includes studies \cite{havasi2022addressing,kim2023probabilistic,keser2023interpretable,sawada2022concept,sheth2023auxiliary,li2024text,hu2024editable}. In the graph concept field, \cite{magister2021gcexplainer} provides a global interpretation of graph neural networks (GNNs) by mapping graphs into a concept space through clustering and offering a human-in-the-loop evaluation. \cite{xuanyuan2023global, Barbiero2023interpretable} extend this approach by incorporating both global and local explanations. For local explanations, they define a concept set, with each neuron represented as a vector with Boolean values indicating concept activation. However, existing works rarely consider semi-supervised settings, which are practical in real-world applications. Our framework addresses these issues effectively.

\noindent {\bf Semi-supervised Learning.}
Semi-supervised learning (SSL) combines the two main tasks of machine learning: supervised learning and unsupervised learning \cite{van2020survey}. It is typically applied in scenarios where labeled data is scarce. Examples include computer-aided diagnosis \cite{zhang2015semi, 8651980}, medical image analysis \cite{HUYNH2022106628, CHEPLYGINA2019280, FILIPOVYCH20111109}, and drug discovery \cite{chen2016nllss}. In these cases, collecting detailed annotated data by experts requires considerable time and effort. However, under the assumption of data distribution, unlabeled data can also assist in building better classifiers \cite{van2020survey}. SSL, also known as self-labeling or self-teaching in its earliest forms, involves the model iteratively labeling a portion of unlabeled data and adding it to the training set for the next round of training \cite{ouali2020overview}. The expectation-maximization (EM) algorithm proposed by \cite{moon1996expectation} uses both labeled and unlabeled data to produce maximum likelihood estimates of parameters. 
\cite{laine2016temporal} and \cite{tarvainen2017mean} focus on consistency regularization. \uppercase\expandafter{\romannumeral2}-model \cite{laine2016temporal} combines both supervised cross-entropy loss and unsupervised consistency loss while perturbing the model and data based on the consistency constraint assumption. A temporal ensembling model integrates predictions from models at various time points. Mean teacher \cite{tarvainen2017mean} addresses the slow updating issue of the temporal ensembling model on large datasets by averaging model weights instead of predicting labels. MixMatch \cite{NEURIPS2019_1cd138d0} unifies and refines the previous approaches of consistency regularization, entropy minimization, and traditional regularization into a single loss function, achieving excellent results. Pseudo labeling, as an effective tool for reducing the entropy of unlabeled data \cite{lee2013pseudo}, has been increasingly attracting the attention of researchers in the field of semi-supervised learning. \cite{arazo2020pseudo} proposes that directly using the model's predictions as pseudo-labels can achieve good results. FixMatch \cite{sohn2020fixmatch} predicts and retains the model, generating high-confidence pseudo-labels. \cite{pham2021meta} continuously adjusts the teacher based on feedback from the student, thereby generating better pseudo-labels. While there has been a plethora of work in the semi-supervised learning field, the focus on semi-supervised concept bottleneck models remains largely unexplored. Our work focuses on this new area.
\section{Preliminaries} \label{sec:pre}
\noindent {\bf Concept Bottleneck Models \cite{koh2020concept}. }
 We consider a classification task with a concept set denoted as $\mathcal{C} =\{p_1, \cdots, p_k\}$ with each $p_i$ being a concept given by experts or LLMs, and a training dataset represented $\mathcal{D} = \{(x^{(i)}, y^{(i)}, c^{(i)})\}_{i=1}^{N}$. Here, for $i\in [N]$, $x^{(i)} \in \mathcal{X} \subseteq \mathbb{R}^d$ represents the feature vector (e.g., an image’s pixels), $y^{(i)} \in \mathcal{Y} \subseteq \mathbb{R}^{l}$ denotes the label ($l$ is the number of classes), ${c}^{(i)} =(c_i^1, \cdots, c_i^k) \in \mathbb{R}^k$ represents the concept vector (a binary vector of length $k$, where each value indicates whether the input $x^{(i)}$ contains that concept). 
In CBMs, the goal is to learn two representations: one called a concept encoder that transforms the input space into the concept space, denoted as $g: \mathbb{R}^d \to \mathbb{R}^k$, and another called label predictor that maps the concept space to the downstream prediction space, denoted as $f: \mathbb{R}^k \to \mathbb{R}^{l}$. Usually, the map $f$ is linear. For any input $x$, we aim to ensure that its predicted concept vector $\hat{c}=g(x)$ and prediction $\hat{y}=f(g(x))$ are close to their underlying counterparts, thus capturing the essence of the original CBMs.

\noindent {\bf Concept Embedding Models \cite{espinosa2022concept}.} As the original CBM is based solely on concept features to determine the predictions of the model, compared to canonical deep neural networks, it will degrade prediction performance. To further improve the performance of CBMs, CEM is developed by \cite{espinosa2022concept}. It achieves this by learning interpretable high-dimensional concept representations (i.e., concept embeddings), thus maintaining high task accuracy while obtaining concept representations that contain meaningful semantic information. For CEMs, we use the same setting as that of \cite{espinosa2022concept, ismail2023concept}. For each input $x$, the concept encoder learns $k$ concept representations $\hat{c}_1, \hat{c}_2, \ldots, \hat{c}_k$, each corresponding to one of the $k$ ground truth concepts in the training dataset. In CEMs, each concept $c_i$ is represented using two embeddings $\bm{\hat{c}}_{i}^{+},\bm{\hat{c}}_{i}^{-} \in \mathbb{R}^{m}$, each with specific semantics, i.e., the concept is \textit{True} (activate state) and concept is \textit{False} (negative state), where hyper-parameter $m$ is the embedding dimension. We use a DNN $\psi(x)$ to learn a latent representation $\bm{h} \in \mathbb{R}^{n_{h}}$, which is used as input to the CEM embedding generator, where $n_{h}$ is the dimension of the latent representation. The CEM embedding generator $\phi$ feeds $\bm{h}$ into two fully connected concept-specific layers to learn two concept embeddings in $\mathbb{R}^{m}$.
\begin{equation*}
    \bm{\hat{c}}_{i}=\phi _{i}\left( \boldsymbol{h} \right) =a\left(W_{i}\boldsymbol{h}+\boldsymbol{b}_{i} \right).
\end{equation*}
Then we use a differential scoring function $s:\mathbb{R}^{2m}\rightarrow \left[ 0,1 \right]$, to achieve the alignment of concept embeddings $\bm{\hat{c}}_{i}^{+},\bm{\hat{c}}_{i}^{-}$ and ground-truth concepts $c_i$. It is trained to predict the probability $\hat{p}_i:=s\left( \left[ \boldsymbol{\hat{c}}_{i}^{+},\boldsymbol{\hat{c}}_{i}^{-} \right] ^{\top} \right) =\sigma \left( W_s\left[ \left[ \boldsymbol{\hat{c}}_{i}^{+},\boldsymbol{\hat{c}}_{i}^{-} \right] ^{\top} \right] +\boldsymbol{b}_s \right) $ that the concept $c_i$ is active in the embedding space. We get the final concept embedding $\bm{\hat{c}_{i}}$, as follows:
\begin{equation*}
    \bm{\hat{c}}_{i}:=\hat{p}_i\hat{c}_{i}^{+}+(1-\hat{p}_i)\hat{c}_{i}^{-}.
\end{equation*} 
At this point, we understand that we can obtain high-quality concept embeddings rich in semantics through CEMs. In the subsequent section \ref{sec:method}, we will effectively utilize these representations of concepts and further optimize their interpretability through our proposed framework SSCBM. %Additionally, We will align it with the input saliency map to address this issue affecting the interpretability of CBMs.

\noindent {\bf Semi-supervised Setting.}
Now, we consider the setting of semi-supervised learning for concept bottleneck models. As mentioned earlier, a typical training dataset for CBMs can be represented as $\mathcal{D} = \{(x^{(i)}, y^{(i)}, c^{(i)})\}_{i=1}^{N}$, where $x^{(i)} \in \mathcal{X}$ represents the input feature. However, in semi-supervised learning tasks, the set of feature vectors typically consists of two parts, $\mathcal{X} = \{\mathcal{X}_{L}, \mathcal{X}_{U}\}$, where $\mathcal{X}_{L}$ represents a small subset of labeled data and $\mathcal{X}_{U}$ represents the remaining unlabeled data. Generally we assume $|\mathcal{X}_{L}| \ll |\mathcal{X}_{U}|$. We assume that $x^{(j)} \in \mathcal{X}_{L}$ is labeled with a concept vector $c^{(j)}$ and a label $y^{(j)}$, and for any $x^{(i)} \in \mathcal{X}$, there only exists a corresponding label $y^{(i)} \in \mathcal{Y}$. Note that our method can be directly extended to the fully semi-supervised case where even the classification labels for feature vectors in $\mathcal{X}_{U}$ are unknown. 

Under these settings, given a training dataset $\mathcal{D} = \mathcal{D}_{L} \cup \mathcal{D}_{U}$  that includes both labeled and unlabeled data, the goal is to train a CBM using both the labeled data $\mathcal{D}_{L}$ and unlabeled data $\mathcal{D}_{U}$. This aims to get better mappings $g: \mathbb{R}^d \to \mathbb{R}^k$ and $f: \mathbb{R}^k \to \mathbb{R}^{l}$ than those trained by using only labeled data, ultimately achieving higher task accuracy and its corresponding concept-based explanation.
\section{Semi-supervised Concept Bottleneck Models}\label{sec:method}
\begin{figure*}[t]
    \centering
    \includegraphics[width=0.9\linewidth]{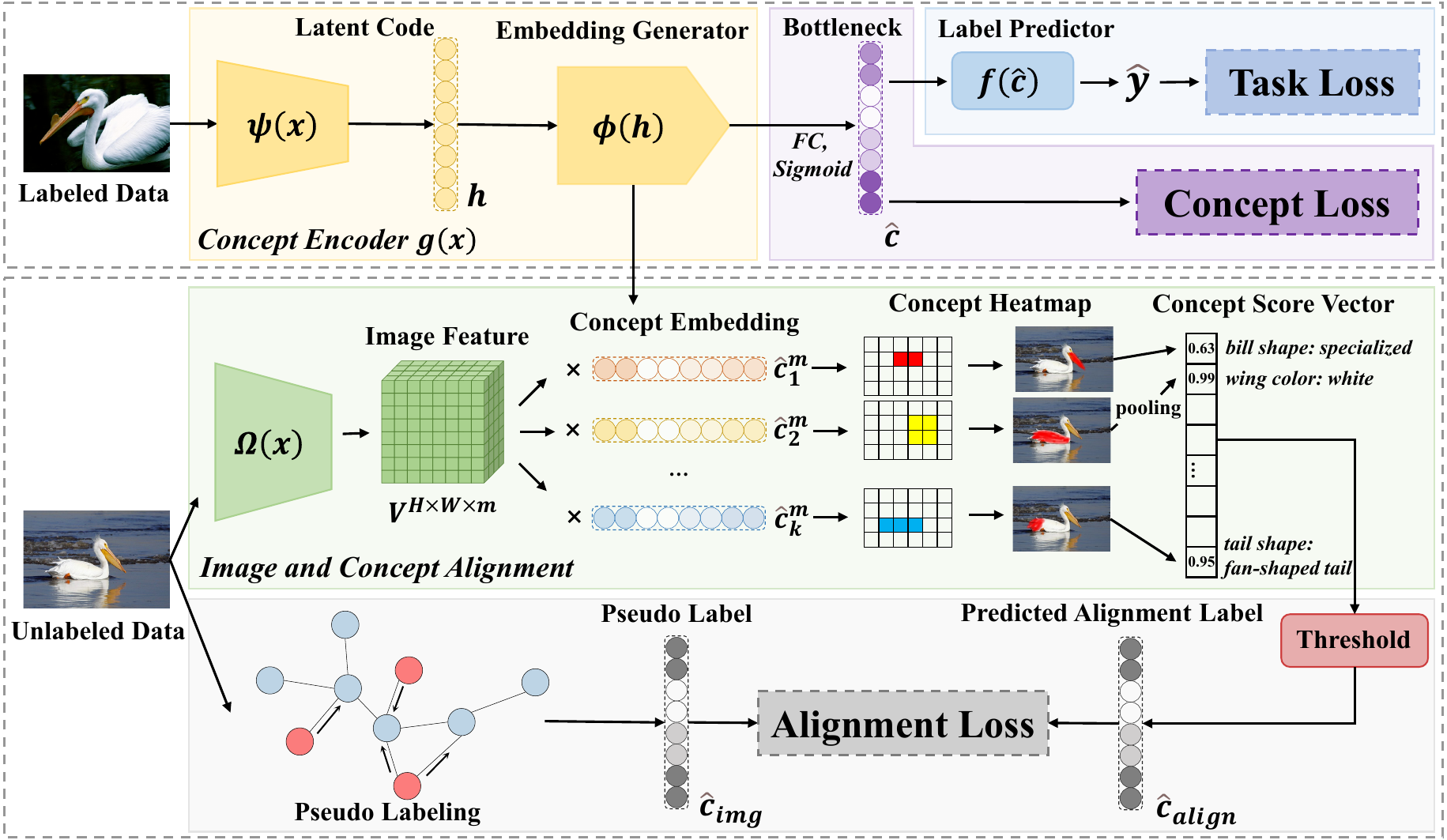}
    \caption{Overall framework of our proposed SSCBM. 
    \label{fig:framework}}
\end{figure*}

In this section, we will elaborate on the details of the proposed SSCBM framework, which is shown in Figure \ref{fig:framework}. SSCBM follows the main idea of CEM.  Specifically, to learn a good concept encoder,  we use different processing methods for labeled and unlabeled data. The labeled data first passes through a feature extractor $\psi$ to be transformed into a latent representation $\bm{h}$, which then enters the concept embedding extractor to obtain the concept embeddings and the predicted concept vector $\bm{\hat{c}}$ for the labeled data, which is compared to the ground truth concept to compute the concept loss. Moreover, the label predictor predicts $\bm{\hat{y}}$ based on $\bm{\hat{c}}$, and calculates the task loss.

For unlabeled data, we first extract image features $V$ from the input using an image encoder. Then, we use the KNN algorithm to assign a pseudo-label $\bm{\hat{c}_{img}}$ to each unlabeled data, which has been experimentally proven to be simple and effective.  In the second step, we compute a heatmap between concept embeddings and the input. After applying a threshold, we obtain the predicted alignment label $\bm{\hat{c}_{align}}$. Finally, we compute the alignment loss between $\bm{\hat{c}_{img}}$ and $\bm{\hat{c}_{align}}$. During each training epoch, we simultaneously compute these losses and update the model parameters based on the gradients.

\subsection{Label Anchor: Concept Embedding Encoder}\label{sec:label}
\noindent {\bf Concept Embeddings.} 
As described in Section \ref{sec:pre}, we obtain high-dimensional concept representations with meaningful semantics based on CEMs. Thus, our concept encoder should extract useful information from both labeled and unlabeled data.

For labeled training data $\mathcal{D}_{L} = \{(x^{(i)}, y^{(i)}, c^{(i)})\}_{i=1}^{|\mathcal{D}_{L}|}$, we follow the original CEM~\cite{espinosa2022concept}, i.e., using a backbone network (e.g., ResNet50) to extract features $\boldsymbol{h} = \{ \psi(x^{(i)}) \}_{i=1}^{|\mathcal{D}_{L}|}$. Then, for each feature, it passes through an embedding generator to obtain concept embeddings $\bm{\hat{c}}_i \in \mathbb{R}^{m \times k}$ for $i \in [k]$. After passing through fully connected layers and activation layers, we obtain the predicted binary concept vector $\bm{\hat{c}} \in \mathbb{R}^{k}$ for the labeled data. The specific process can be represented by the following expression:
\begin{equation*} \boldsymbol{ h^{(j)}} = \psi(x^{(j)}),\quad j = 1, \ldots, |\mathcal{D}_{L}|,
\end{equation*}
\begin{equation*} \boldsymbol{\hat{c}}^{(j)}_{i} = \sigma(\phi(\boldsymbol{h^{(j)}})), \quad i = 1, \ldots, k, \quad j = 1, \ldots, |\mathcal{D}_{L}|,
\end{equation*}

where $\psi$, $\phi$ and $\sigma$ represent the backbone, embedding generator, and activation function, respectively. 

To enhance the interpretability of concept embeddings, we calculate the concept loss utilize binary cross-entropy to optimize the accuracy of concept predictions by computing $\mathcal{L}_c$ based on predicted binary concept vector $\bm{\hat{c}}$ and ground truth concept labels $\bm{c}$:
\begin{equation} \label{l_concept_loss}
    \mathcal{L}_{c} = BCE(\boldsymbol{\hat{c}}, \boldsymbol{c}), 
\end{equation}
where $BCE$ is the binary cross entropy loss.

\noindent {\bf Task Loss.} Since our ultimate task is classification, we also need to incorporate the task loss for the final prediction performance. After obtaining the predicted concept $\boldsymbol{\hat{c}}$, we use a label predictor to predict the final class $\boldsymbol{\hat{y}}$. Then we can define the task loss function using the categorical cross-entropy loss to train our classification model as follows:

\begin{equation} \label{l_task_loss}
\mathcal{L}_{task} = CE(\boldsymbol{\hat{y}}, \boldsymbol{y}),
\end{equation}

Note that for the unlabeled data, we can also calculate the task loss since their class labels are known, to make full use of the data.

\subsection{Unlabel Alignment: Image-Textual Semantics Alignment}\label{sec:unlabel}

\noindent {\bf Pseudo Labeling.}
%As mentioned earlier, the concept encoder of CEMs surpasses CBMs classification performance by learning interpretable high-dimensional concept representations that contains meanful semantics. However, the concept encoder is a supervised training method. 
Unlike supervised data, CEM cannot directly extract useful information from unlabeled data, as the concept encoder is a supervised training architecture. Thus, in practical situations lacking labeled data, one direct approach is to get high-quality pseudo concept labels to train the model. In the following, we will introduce the method we use to obtain pseudo-concept labels.

Firstly, we can naturally think of measuring the similarity between images by calculating the distance in the cosine space. Based on this idea, we can assign pseudo labels to unlabeled data by finding labeled data with similar image features to them.
% For each unlabeled data point, we find the k labeled data points that are closest in cosine space. We then use the inverse of the normalized distances as weights to compute the weighted pseudo concept labels $\boldsymbol{\tilde{c}}$. Inspired by \cite{arazo2020pseudo}, we can obtain pseudo concept labels through the following steps. We initially train the model on a small subset of labeled data, then iteratively predict pseudo labels for each unlabeled data point. We retain only the high-confidence pseudo labels until all data points have pseudo labels. However, unlike typical deep learning models used for classification, concept bottleneck models require learning a mapping from input to concept before predicting labels based on concepts. In the absence of rich concept annotations, the model's performance can significantly decrease, as we have confirmed in practice. On the other hand, we found that simply using the k-nearest neighbors algorithm for pseudo labeling can yield good results. 
Specifically, for each unlabeled training data $x \in \mathcal{D}_{U} = \{(x^{(i)}, y^{(i)})\}_{i=1}^{|\mathcal{D}_{U}|}$, we calculate its cosine distance with all labeled data points $x^{(j)} \in \mathcal{D}_{L} = \{(x^{(j)}, y^{(j)}, c^{(j)})\}_{j=1}^{|\mathcal{D}_{L}|}$ and select the $k$ samples with the smallest distances:
\begin{equation*}
\operatorname{dist}(x, x^{(j)}) = 1- \frac{ \Omega(x) \cdot \Omega(x^{(j)})}{\|\Omega(x)\|_{2} \cdot \|\Omega(x^{(j)})\|_{2}}, \quad j = 1, \ldots, |\mathcal{D}_{L}|,
\end{equation*}
where $\Omega$ is a visual encoders. 
Then, we normalize the reciprocal of the cosine distance between the nearest $k$ data points and $x$ as weights. We use these weights to compute a weighted average of the concept labels of these $k$ data points, obtaining the pseudo concept label for $x$. In this way, we can obtain pseudo concept labels $\bm{\hat{c}_{img}}$ for all $x \in \mathcal{D}_{U}$.

In our experiments, we find that directly feeding pseudo-concept labels generated by KNN to CEM has satisfactory performance.  However, this simple labeling method can lead to alignment issues in the concept embeddings learned by the CEM's concept encoder. Specifically, the predicted concepts $\bm{\hat{c}}$ could not have a relation to the corresponding features $V$ in the image, which hinders the effectiveness of CEM as a reliable interpretability tool. Moreover, due to misalignment, this will also degrade the prediction performance of the concept encoder.  In the following, we aim to address such a misalignment issue. 

\noindent {\bf Generating Concept Heatmaps.}  Our above pseudo concept labels via KNN have already contained useful information for prediction. Thus, our goal is to provide these labels with more information about their relation to the corresponding features. To achieve this, we first get another pseudo-concept label via these relations, which are calculated by the similarity between the concept embedding and the input image, namely concept heatmaps. Specifically, given an image $x$, we first have its feature map $V \in \mathbb{R}^{H \times W \times m}$ for each concept extracted by $V = \Omega(x)$, where $H$ and $W$ are the height and width of the feature map. 

Given $V$ and the $i$-th concept embedding $\boldsymbol{c}_i$, we can obtain a heatmap $\mathcal{H}_i$, i.e., a similarity matrix that measures the similarity between the concept and the image can be obtained by computing their cosine distance:
\begin{equation*}
\mathcal{H}_{p,q,i} = \frac{\boldsymbol{{\hat{c}}_i^m}\cdot  V_{p,q}}{||\boldsymbol{\hat{c}_i^m}||\cdot ||V_{p,q}||},\quad p = 1, \ldots , H, \quad q = 1, \ldots , W
\end{equation*}
where $p,q$ are the $p$-th and $q$-th positions in the heatmaps, and $\mathcal{H}_{p,q,i}$ represents a local similarity score between $V$ and $\boldsymbol{\hat{c}}_i^m$. Intuitively, $\mathcal{H}_{i}$ represents the relation of each part of the image with the $i$-the concept. Then, we derive heatmaps for all concepts, denoted as $\{\mathcal{H}_1, \mathcal{H}_2, \ldots, \mathcal{H}_k\}$.

\noindent {\bf Calculating Concept Scores and Concept Labels.}
As average pooling performs better in downstream classification tasks \cite{yan2023robust}, we apply average pooling to the heatmaps to deduce the connection between the image and concepts: $s_i = \frac{1}{P \cdot Q} \sum_{p=1}^{P} \sum_{q=1}^{Q} \mathcal{H}_{p,q,i}$. Intuitively, $s_i$ is the refined similarity score between the image and concept ${\hat{c}}_i^m$. Thus, a concept vector $\boldsymbol{s}$ can be obtained, representing the similarity between an image input $x$ and the set of concepts: $\boldsymbol{s} = (s_1, \ldots , s_k)^\top$. $\boldsymbol{s}$ can be considered as a soft concept label which is got by similarity. Next, we have to transform it into a hard concept label $\bm{\hat{c}_{align}}$.  we determine the presence of a concept attribute in an image based on a threshold value derived from an experiment. If the value $s_i$ exceeds this threshold, we consider the image to possess that specific concept attribute and set the concept label to be \textit{True}. We can obtain predicted concept labels for all unlabeled data.

\noindent {\bf Alignment of Image.}
%To align images with concept labels, we determine the presence of a concept attribute in an image based on a threshold value derived from an experiment. If the value $s_i$ exceeds this threshold, we consider the image to possess that specific concept attribute and set the concept label to be \textit{True}. We can obtain predicted concept labels for all images $ \hat{c} = \{C_1, \ldots, C_{|\mathcal{D}_{U}|}\} $, where $C_i\in \{0, 1\}^k$ is the concept label for the $i$-th sample. To ensure the truthfulness of concepts, we discard all concepts for which the similarity across all images is below 0.6. 
Based on our earlier discussions, on the one hand, the concept encoder should learn information from $\bm{\hat{c}_{img}}$. On the other hand, it should also get concept embeddings which can get good similarity-based concept labels $\bm{\hat{c}_{align}}$ for alignment with the input image. Thus, we need a further alignment loss to achieve these two goals.  Specifically, we compute the alignment loss as follows:
\begin{equation} \label{l_alignment_loss}
    \mathcal{L}_{align} = BCE(\bm{\hat{c}_{img}}, \bm{\hat{c}_{align}}). 
\end{equation}

\subsection{Final Objective}
In this section, we will discuss how we derive the process of network optimization. First, we have the concept loss $\mathcal{L}_{c}$ in (\ref{l_concept_loss}) to improve the interpretability of the concept embeddings. Also, since the concept embeddings are entered into the label predictor to output the final prediction, we also have a task loss between the predictions given by the concept bottleneck and ground truth, which is shown in (\ref{l_task_loss}). In the context of binary classification tasks, we employ binary cross-entropy (BCE) as our loss function. For multi-class classification tasks, we use cross-entropy as the measure. Finally, to align the images with the concept labels, we computed the loss of alignment in (\ref{l_alignment_loss}).
Formally, the overall loss function of our approach can be
formulated as:
\begin{equation} \label{l_final}
\mathcal{L} = \mathcal{L}_{task} + \lambda_1 \cdot \mathcal{L}_{c} + \lambda_2 \cdot \mathcal{L}_{align},   
\end{equation}
where $\lambda_1, \lambda_2$ are hyperparameters for the trade-off between interpretability and accuracy.
\section{Experiments}\label{sec:experiment}

In this section, we will conduct experimental studies on the performance of our framework. Specifically, we evaluate the utility of our method according to the concept and prediction of the class label. We also design interpretability evaluation and test-time intervention to verify the interpretability and alignment performance of our method. Finally, we give the ablation study to verify why we need the alignment loss. Details and additional results are in Appendix \ref{App:Setup}. 

\begin{table}[ht]
\centering
\caption{Concept and task accuracy results of SSCBM at a labeled ratio of 0.1, compared to baselines in the fully supervised setting.\label{tab:perform}}
\vspace{-6pt}

\setlength{\tabcolsep}{2pt}

\resizebox{\linewidth}{!}{
\begin{tabular}{ccccccccc}
\toprule
\multirow{2}{*}{\textbf{Method}} & \multicolumn{2}{c}{\textbf{CUB}} & \multicolumn{2}{c}{\textbf{AwA2}}&
\multicolumn{2}{c}{\textbf{WBCatt}}&
\multicolumn{2}{c}{\textbf{7-point}}\\
\cmidrule(lr){2-3} \cmidrule(lr){4-5} \cmidrule(lr){6-7} \cmidrule(lr){8-9}
 & \textbf{Concept} & \textbf{Task} & \textbf{Concept} & \textbf{Task} & \textbf{Concept} & \textbf{Task} & \textbf{Concept} & \textbf{Task} \\
\midrule
CBM & 93.99\% & 67.33\% & 96.48\% & 88.71\% & 94.18\% & 99.71\% & 74.34\% & 75.44\% \\
CEM & 96.39\% & 79.82\% & 95.91\% & 87.00\% & 95.33\% & 99.71\% & 77.15\% & 75.85\% \\

% \rowcolor{gray!25}

SSCBM & 90.88\% & 67.67\% & 96.48\% & 89.77\% & 93.98\% & 99.68\% & 73.67\% & 70.09\% \\

\bottomrule
\end{tabular}}
\vspace{-12pt}
\end{table}
\begin{table*}[ht]
\centering
\caption{Results of concept and task accuracy for different datasets with different portions of labeled data. Since the label-free CBM \cite{oikarinen2023label} does not use the concept set included in the dataset but instead relies on GPT-3 to generate concepts based on class names, we do not report the concept accuracy here. \label{tab:ratio}}
\vspace{-6pt}
\resizebox{0.75\linewidth}{!}{
\begin{tabular}{cccccccccc}
\toprule
\multirow{2}{*}{\textbf{Dataset}} & \multirow{2}{*}{\textbf{Labeled Ratio}} & \multicolumn{2}{c}{\textbf{CBM+SSL}} & \multicolumn{2}{c}{\textbf{CEM+SSL}} & \multicolumn{2}{c}{\textbf{Label-free CBM}} & \multicolumn{2}{c}{\textbf{SSCBM}} \\
\cmidrule(lr){3-4} \cmidrule(lr){5-6} \cmidrule(lr){7-8} \cmidrule(lr){9-10}
 & & \textbf{Concept} & \textbf{Task} & \textbf{Concept} & \textbf{Task} & \textbf{Concept} & \textbf{Task} & \textbf{Concept} & \textbf{Task} \\
\midrule
\multirow{5}{*}{CUB} & K=1 & 83.11\% & 5.51\% & 82.36\% & 59.35\% & - & 74.31\% & 88.99\% & 66.72\% \\
& 0.05 (K=2) & 84.51\% & 8.35\% & 83.72\% & 62.20\% & - & 74.31\% & 90.04\% & 67.43\% \\
& 0.1 (K=3) & 84.96\% & 9.84\% & 84.03\% & 63.12\% & - & 74.31\% & 90.88\% & 67.67\% \\
& 0.15 (K=4) & 85.47\% & 9.96\% & 84.30\% & 64.14\% & - & 74.31\% & 91.47\% & 68.36\% \\
& 0.2 (K=5) & 86.67\% & 16.43\%& 86.83\% &67.64\% & -& 74.31\% & 92.09 \% & 70.07\% \\

% \midrule
% \multirow{5}{*}{CelebA} & K=1 & - &- & - & - & - & - & &  \\
% & 0.05 (K=1) & - & - & -& - & - & - &  &  \\
% & 0.1 (K=) &- & - & - & - & - & - &  &  \\
% & 0.15 (K=) & - & -& -& - & - & - &  &  \\
% & 0.2 (K=) & -& - & - & - & - & - &  &  \\

\midrule
\multirow{5}{*}{AWA2} & K=1 & 71.32\% & 90.87\% & 65.55\% & 84.44\% & - &73.67\% & 85.05\% & 89.02\% \\
& 0.05 (K=30) & 77.23\% & 91.11\% & 68.31\% & 85.01\% & - &73.67\% & 95.42\% & 89.37\% \\
& 0.1 (K=60)& 80.45\% & 91.03\% & 69.96\% & 85.48\% & - & 73.67\% & 96.48\% & 89.77\% \\
& 0.15 (K=90) & 83.46\% & 91.00\% & 72.14\% & 86.61\% & - &73.67\% & 96.81\% & 90.54\% \\
& 0.2 (K=120)& 85.76\% & 91.12\% & 72.14\% & 86.61\% & - & 73.67\% & 96.81\% & 90.44\% \\

\midrule
\multirow{5}{*}{WBCatt} & K=1 &79.06\% &99.39\%&70.27\%&98.64\%&-&41.92\%&91.48\%&99.13\%\\
& 0.05 (K=62)&81.08\% &99.48\%&73.82\%&99.52\%&-&41.92\%&93.53\%&99.61\%\\
& 0.1 (K=124) & 85.48\%&99.32\%&72.25\%&99.29\%&-&41.92\%&93.98\%&99.68\%\\
& 0.15 (K=186)&85.39\% &99.68\%&72.68\%&99.58\%&-&41.92\%&94.42\%&99.71\%\\
& 0.2 (K=247) &87.07\% &99.74\%&74.14\%&99.52\%&-&41.92\%&94.42\%&99.71\%\\
\midrule
\multirow{5}{*}{7-point} & K=1 &59.91\%&55.95\%&62.78\%&66.09\%&-&55.44\%&66.58\%&66.84\%\\
& 0.05 (K=5)&65.36\%&57.47\%&67.85\%&67.09\%&-&55.44\%&70.98\%&68.77\%\\
& 0.1 (K=9) &68.82\%&55.70\%&72.23\%&66.33\%&-&55.44\%&73.67\%&70.09\%\\
& 0.15 (K=13)&66.14\%&59.75\%&66.54\%&67.09\%&-&55.44\%&73.94\%&72.56\%\\
& 0.2 (K=17) &70.29\%&60.44\%&73.04\%&66.84\%&-&55.44\%&76.52\%&74.56\%\\
\bottomrule

\end{tabular}}

\vspace{-4pt}

\end{table*}

\subsection{Experimental Settings}
\noindent{\bf Datasets.}
We evaluate our methods on four real-world image tasks: \textit{CUB}\cite{he2019fine}, 
% \textit{CelebA}\cite{liu2015faceattributes}, 
\textit{AwA2}\cite{niu2018learning}, \textit{WBCatt}\cite{tsutsui2023wbcatt} and \textit{7-point}\cite{Kawahara2018-7pt}. See Appendix \ref{app:dataset} for a detailed introduction.

\noindent{\bf Baselines.}
We compare our SSCBM with Concept Bottleneck Model (CBM) and Concept Embedding Model (CEM) mentioned in Section \ref{sec:pre}. Since those baselines do not inherently include settings for semi-supervised learning, to ensure fairness in the evaluation, we use the same pseudo-labeling approach as SSCBM. We utilize the KNN algorithm to annotate unlabeled data based on labeled data, which is mentioned in Section \ref{sec:unlabel}.  For CBM~\cite{koh2020concept}, we adopt the same setting and architecture as in the original CBM. And for CEM, we follow the same setting as in ~\cite{espinosa2022concept}. We also compare label-free CBM \cite{oikarinen2023label}, which is an unsupervised CBM method.
 
\noindent{\bf Evaluation Metrics.} To evaluate the utility, we consider the accuracy for the prediction of class and concept labels. Specifically, \textit{concept accuracy} measures the model's prediction accuracy for concepts:
    $\mathcal{C}_{acc} = \frac{1}{N} \cdot \frac{1}{k} \sum_{i=1}^N \sum_{j=1}^k \mathbb{I}(\hat{c}_{j}^{(i)}=c_{j}^{(i)}). $ 
\textit{Task accuracy} measures the model's performance in predicting downstream task classes:
    $\mathcal{A}_{acc} = \frac{1}{N} \sum_{i=1}^N \mathbb{I}(\bm{\hat{y}}^{(i)}=\bm{y}^{(i)}).$

To evaluate the interpretability, besides the concept accuracy (due to the structure of CBM, concept accuracy can also reveal the interpretability), similar to previous work~\cite{koh2020concept,espinosa2022concept}, we will show \textit{visualization results}. Moreover, we evaluate the performance of the \textit{test-time intervention}.

\noindent{\bf Implementation Details.}
All experiments are conducted on a Tesla V100S PCIe 32 GB GPU and an Intel Xeon Processor CPU. See Appendix \ref{app:imple_detail} for more details.

\subsection{Evalaution Results on Utility}
To comprehensively evaluate the model's performance in the semi-supervised learning setting, we simulate real-world application scenarios. We consider labeling K samples for each class in the dataset, with an extreme case where K=1. We also conduct experiments with labeled ratios of 0.05, 0.1, 0.15, and 0.2. We also compare the performance of SSCBM with the baseline under the fully supervised setting. The experimental results are shown in Table \ref{tab:perform} and \ref{tab:ratio}. Additionally, we also explored the impact of different backbones on model performance, which can be found in the Appendix \ref{app:backbone}.

Table \ref{tab:perform} shows the performance comparison of SSCBM at a labeled ratio of 0.1 with CEM and CBM under the fully supervised setting. It can be observed that, although CEM performs the best on CUB, WBCatt and 7-point, SSCBM is only slightly inferior to it. On the AwA2 dataset, its performance even surpasses that of CEM under the fully supervised setting, which sufficiently demonstrates the superiority of our method.

In Table \ref{tab:ratio}, it can be observed that as the proportion of labeled data gradually increases, the concept and class prediction accuracy of all models improve. SSCBM outperforms all baselines. On the CUB dataset, when only one labeled sample per class is available, SSCBM achieves 88.99\% concept accuracy and 66.72\% class accuracy, which are 6.63\% and 7.37\% higher, respectively, than the performance achieved by CEM in a semi-supervised learning setting. This demonstrates the effectiveness of the Image-Textual Semantics Alignment module. The 7-point dataset is the most challenging, and on this dataset, SSCBM still achieves the best performance across all settings. As the amount of labeled data increases, the concept prediction accuracy improves from 66.58\% to 76.52\%. We also note that on WBCatt and 7-point, the performance of Label-free CBM is poor. This is mainly because both datasets contain only five classes, which prevents Label-free CBM from fully leveraging the world knowledge of LLMs to generate an effective concept set, resulting in poor performance. This further demonstrates that semi-supervised learning scenarios are more realistic and practical for real-world datasets.

\subsection{Interpretability Evaluation}
% For interpretability, in the previous section, we have shown that the concept accuracy of SSCBM with a small labeled ratio is very close to that of CEM, indicating it inherits CEM's interpretability. 

Besides the concept accuracy, note that in SSCBM, we also have a pseudo label revived by the alignment between the concept embedding and the input saliency map. We use the alignment loss to inherit this alignment. Thus, we will evaluate the alignment performance here to show the faithfulness of the interpretability given by SSCBM. We measure our alignment performance by comparing the correctness of the concept saliency map with the concept semantics in Figure \ref{fig:bird3}. See Appendix \ref{app:imple_detail} for detailed experimental procedures for generating saliency maps. Results show that our concept saliency map matches the concept semantics, indicating the effectiveness of our alignment loss. In Appendix \ref{app:saliency_map}, we also provide our additional interpretability evaluation in  Figure \ref{fig:bird4}-\ref{fig:7point5}.

\begin{figure*}[htbp] 
    \setlength{\tabcolsep}{0.2pt} % Default value: 6pt
    \renewcommand{\arraystretch}{1.0} % Default value: 1
    \begin{center}
    \resizebox{0.85\linewidth}{!}{
    \begin{tabular}
    {@{\extracolsep{\fill}}cccc}
    \begin{tabular}{c}
        \includegraphics[width=0.252\linewidth]{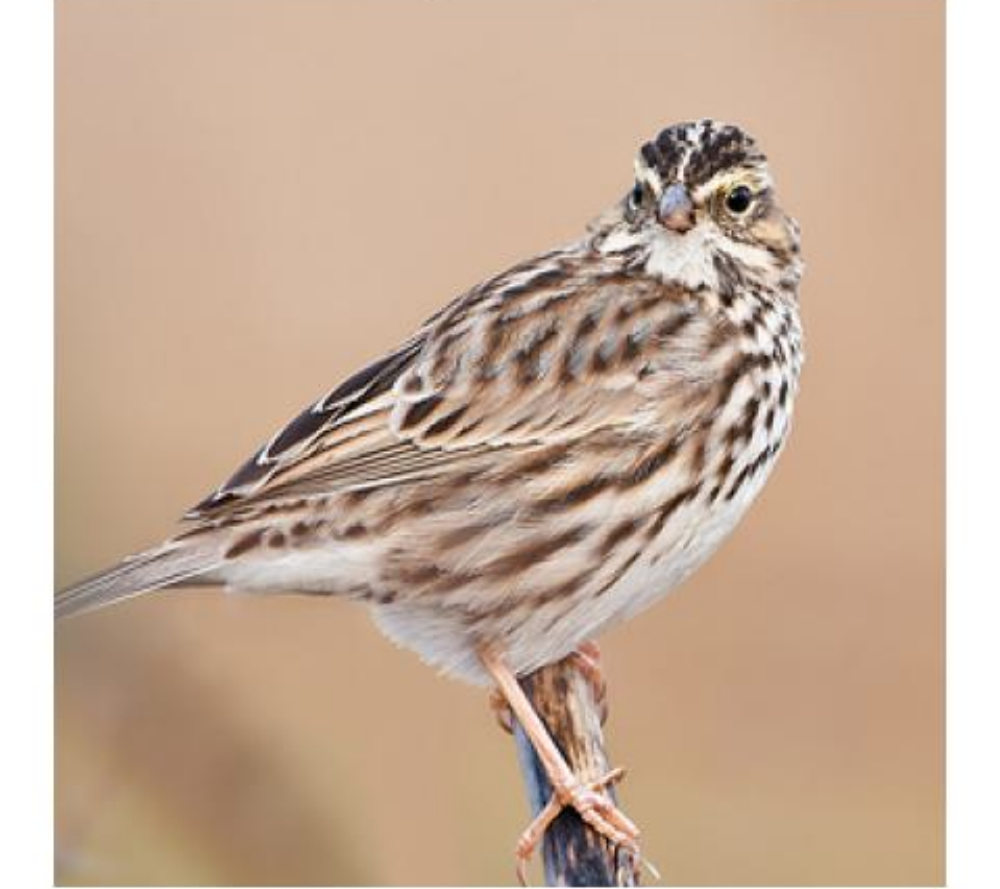} \\
        \footnotesize{(a) Original Image}
    \end{tabular} &
    \begin{tabular}{c}
        \includegraphics[width=0.247\linewidth]{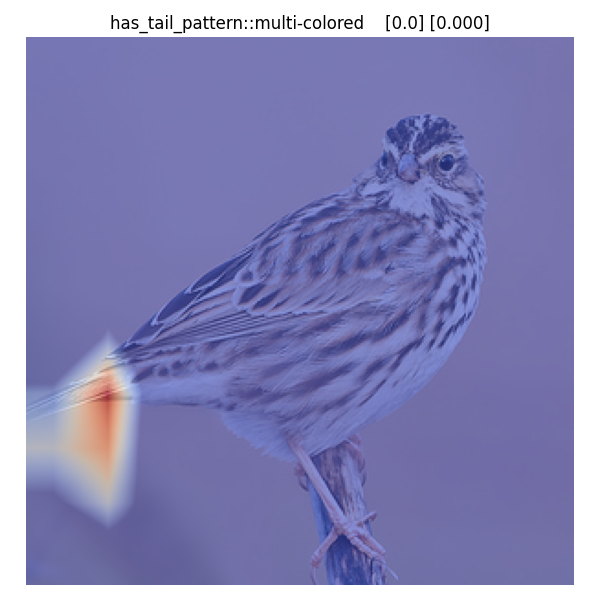} \\
        \footnotesize{(b) Tail Pattern: Multi-colored}
    \end{tabular} &
    \begin{tabular}{c}
        \includegraphics[width=0.247\linewidth]{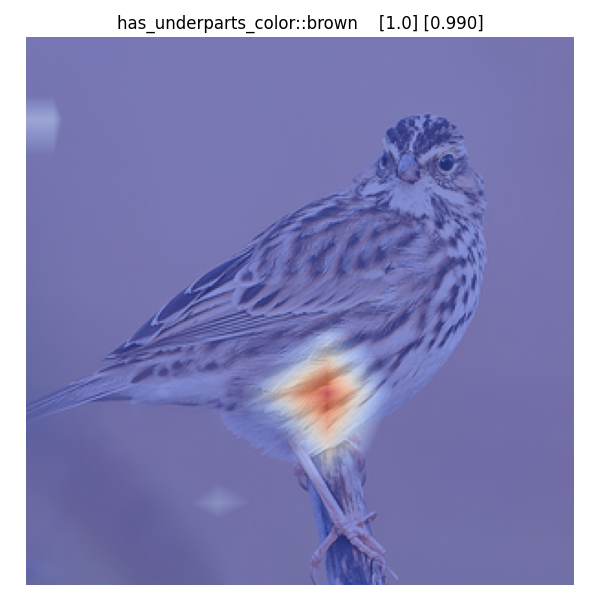} \\
        \footnotesize{(c) Underparts Color: Brown}
    \end{tabular} &
    \begin{tabular}{c}
        \includegraphics[width=0.247\linewidth]{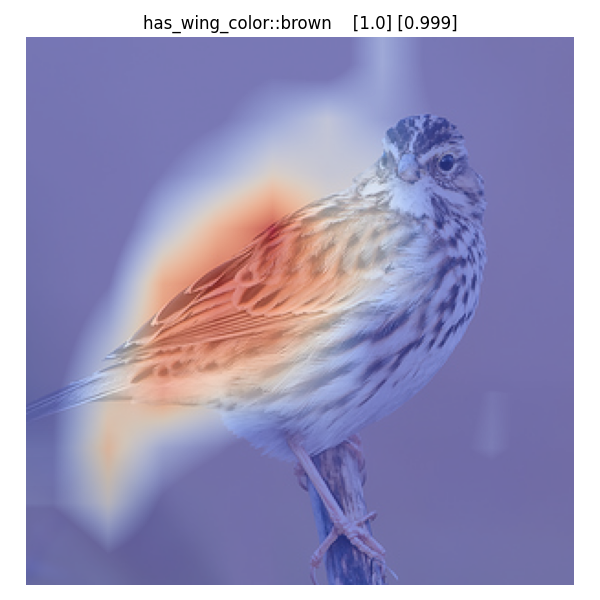} \\
        \footnotesize{(d) Wing Color: Brown}
    \end{tabular} \\  
    \end{tabular}}
    
    \caption{
    The concept saliency map for the CUB dataset (savannah sparrow) demonstrates that our proposed SSCBM achieves meaningful alignment between the ground truth concepts and the input image features. The first image on the left is the original input image. The three images on the right show the aligned regions for different concepts using SSCBM. The text below each image indicates the specific concept, the ground truth concept label, and the prediction result given by SSCBM. \label{fig:bird3}}
    \end{center}

    \vspace{-6pt}

\end{figure*}

% \begin{figure}[htbp]
%     \centering
%     \begin{minipage}[b]{0.37\textwidth}
%         \centering
%         \includegraphics[height=4cm]{source/zhexian_c-l.pdf}
%         \caption{Test-time Intervention Results on CUB Dataset.}
%         \label{fig:image1}
%     \end{minipage}
%     \hfill
%     \begin{minipage}[b]{0.45\textwidth}
%         \centering
%         \includegraphics[height=4cm]{source/演示文稿1.pdf}
%         \caption{An Example of a Successful Intervention.}
%         \label{fig:image2}
%     \end{minipage}
% \vspace{-12pt}
% \end{figure}

% \iffalse
% \begin{figure}[htbp]
%     \centering
% \begin{tabular}{cc}
% \includegraphics[width=0.35\linewidth]{source/interv_cub.pdf}     &  
% \includegraphics[width=0.45\linewidth]{source/example-cub.pdf}
% \\
% \end{tabular}
% \caption{\label{fig:image1} \textbf{Left:} Test-time intervention results on the CUB dataset. \textbf{Right:} An example of successful intervention. \label{fig:sidebyside}}
% \end{figure}
% \fi

\begin{figure*}[htbp]
    \centering
\begin{tabular}{cc}
\includegraphics[width=0.18\linewidth]{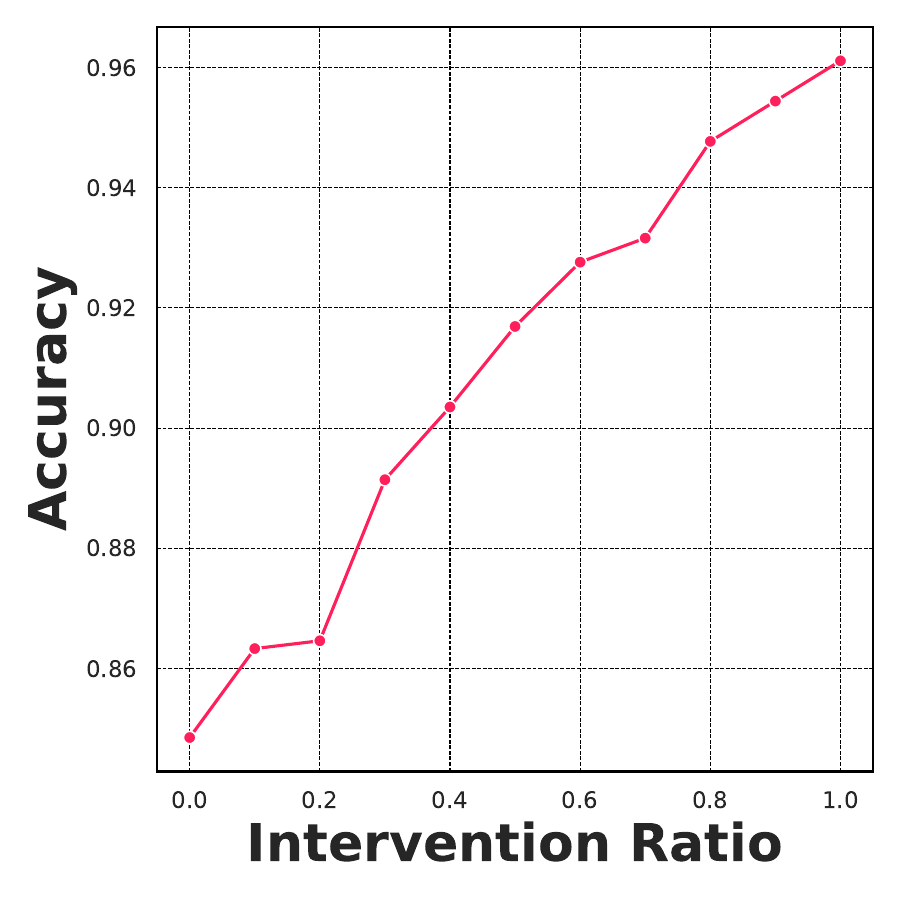}     &  
\includegraphics[width=0.45\linewidth]{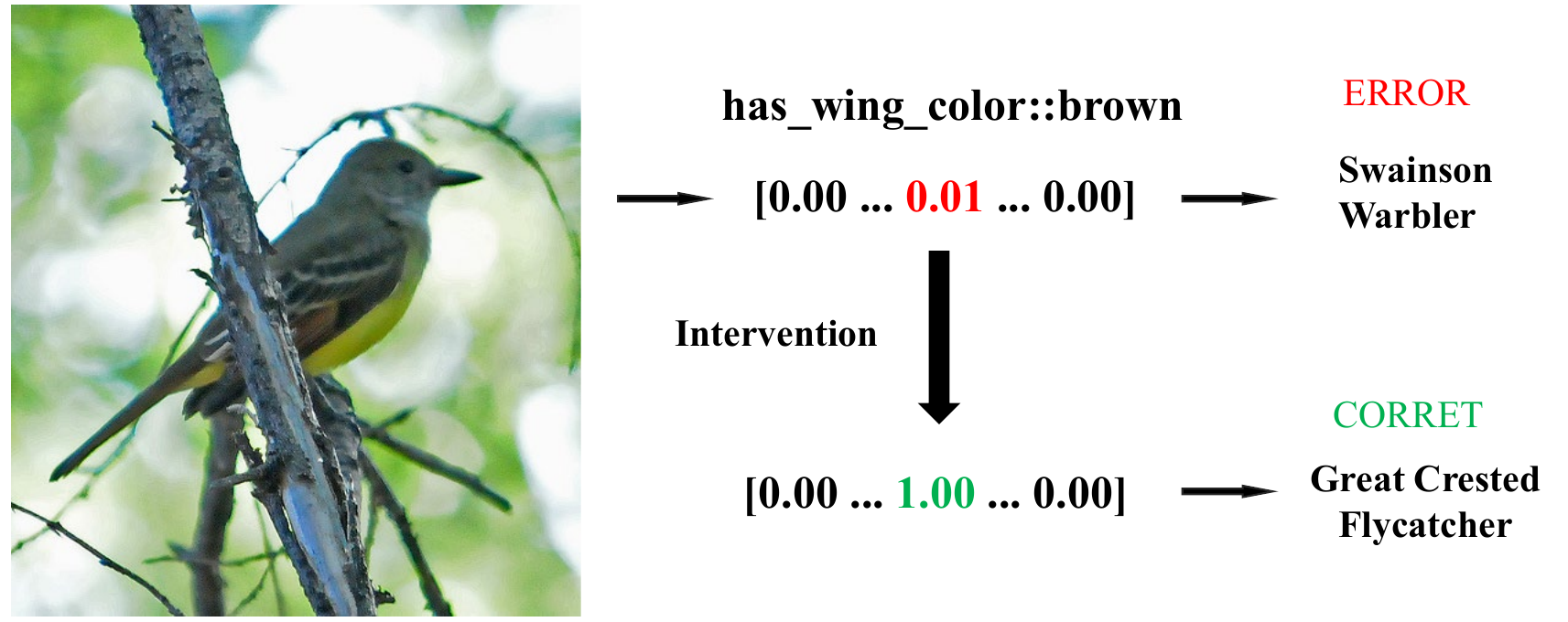} \\
  % (a) & (b)
\end{tabular}
\vspace{-6pt}
\caption{\label{fig:image1} \textbf{Left:} Performance with different ratios of intervened concepts on CUB dataset. \textbf{Right:} An example of successful intervention. \label{fig:sidebyside}}

\end{figure*}

\subsection{Test-time Intervention}
Test-time intervention enables human users to interact with the model at inference time. We test our test-time intervention by correcting the ratio 10\% to 100\% of the concept labels in the concept predictor. We adopt individual intervention for CelebA and AwA2, as there are no grouped concepts. For CUB, we perform the group intervention, i.e., intervene in the concepts with associated attribution. For example, the breast color::yellow, breast color::black, and breast color::white are the same concept group. So, we only need to correct the concept label in the group. We expect that the model performance will steadily increase along with the ratio of concept intervention, indicating that the model learned such correct label information and automatically corrected other labels.  

The results in Figure \ref{fig:image1}  demonstrate the robustness of our model and an increasing trend to learn the information about the concept, indicating our interpretability and model prediction performance. This lies in our alignment loss in effectively learning the correct information pairs in unlabeled and labeled data. The results in Figure \ref{fig:image1}  also show that by changing the wing color to brown, we successfully caused the model to predict the Great Crested Flycatcher instead of the Swainson Warbler. More results are in Appendix \ref{app:testtime}.

\begin{table}[htbp]
\centering

\vspace{-6pt}

\caption{Results of ablation study. \label{tab:ablation}}

\vspace{-6pt}

\setlength{\tabcolsep}{2pt}

\resizebox{\linewidth}{!}{
\begin{tabular}{cccccccc}
\toprule
\multirow{2}{*}{} & \multirow{2}{*}{\textbf{Ratio}} & \multicolumn{2}{c}{\textbf{$w/o$ img}} & \multicolumn{2}{c}{\textbf{$w/o$ align}} &  \multicolumn{2}{c}{\textbf{full model}} \\
\cmidrule(lr){3-4} \cmidrule(lr){5-6} \cmidrule(lr){7-8} 
 & & \textbf{Concept} & \textbf{Task} & \textbf{Concept} & \textbf{Task} & \textbf{Concept} & \textbf{Task} \\
\midrule
\multirow{5}{*}{CUB} 

&0.0001 (K=1) & 87.77\% & 59.89\% & 80.87\% & 65.15\% & 88.99\% & 66.72\% \\
&0.05 (K=2)   & 90.01\% & 65.98\% & 80.60\% & 65.67\% & 90.04\% & 67.43\% \\
&0.1 (K=3)    & 91.02\% & 67.28\% & 80.60\% & 64.69\% & 90.88\% & 67.67\% \\
&0.15 (K=4)   & 90.60\% & 67.97\% & 80.65\% & 65.43\% & 91.47\% & 68.36\% \\
&0.2 (K=5)    & 91.94\% & 68.99\% & 80.73\% & 64.45\% & 92.09\% & 70.07\%\\

\midrule
\multirow{5}{*}{WBCatt} 
&0.0001 (K=1) & 88.44\% & 99.68\% & 57.80\% & 99.51\% & 91.48\% & 99.13\% \\
&0.05 (K=62)  & 92.17\% & 99.29\% & 57.93\% & 99.51\% & 93.53\% & 99.61\% \\
&0.1 (K=124)  & 92.91\% & 99.52\% & 60.49\% & 99.64\% & 93.98\% & 99.68\% \\
&0.15 (K=186) & 93.55\% & 99.58\% & 59.13\% & 99.65\% & 94.42\% & 99.71\% \\
&0.2 (K=247)  & 94.19\% & 99.61\% & 60.65\% & 99.70\% & 94.42\% & 99.71\%\\

\midrule
\multirow{5}{*}{7-point} 
&0.0001 (K=1) & 62.31\% & 64.30\% & 55.07\% & 66.33\% & 66.58\% & 66.84\% \\
&0.05 (K=5)   & 69.42\% & 65.06\% & 56.75\% & 68.10\% & 70.98\% & 68.77\% \\
&0.1 (K=9)    & 70.06\% & 67.34\% & 56.07\% & 65.82\% & 73.67\% & 70.09\% \\
&0.15 (K=13)  & 73.39\% & 68.10\% & 56.23\% & 65.82\% & 73.94\% & 72.56\% \\
&0.2 (K=17)   & 74.44\% & 68.61\% & 56.11\% & 65.57\% & 76.52\% & 74.56\%\\
\bottomrule

\vspace{-28pt}

\end{tabular}}
\end{table}

\subsection{Ablation Study}
We then give a finer study on the two kinds of pseudo-labels to demonstrate that each one is indispensable in bolstering the efficacy of SSCBM. Specifically, based on our method in Section \ref{sec:unlabel}, we have three types of pseudo labels available: pseudo concept label $\bm{\hat{c}_{img}}$, $\bm{\hat{c}_{align}}$, and label $\bm{\hat{c}}$ predicted by the concept embedding. We first remove $\bm{\hat{c}_{img}}$ and calculate our alignment loss using $\bm{\hat{c}_{align}}$ and $\bm{\hat{c}}$; conversely, we remove  $\bm{\hat{c}_{align}}$ and calculate the alignment loss using $\bm{\hat{c}_{img}}$ and $\bm{\hat{c}}$. 

From the results presented in Table \ref{tab:ablation}, it is evident that removing the $\bm{\hat{c}_{img}}$ component significantly degrades the performance of SSCBM at both the concept level and class level. This indicates that the pseudo-concept labels via KNN contain necessary information about the ground truth. The concept encoder needs to extract information from such labels to get better performance. This situation is similar when removing the $\bm{\hat{c}_{align}}$ component, indicating that aligning the concept embedding and the input saliency map can further extract useful information from the input image and, thus, is beneficial to improve the performance. Our observations underscore the high degree of joint effectiveness of two kinds of pseudo-concept labels within our objective function, collectively contributing to the enhancement of model prediction and concept label prediction.
\section{Conclusion}
\vspace{-6pt}
The training of current CBMs heavily relies on the accuracy and richness of annotated concepts in the dataset. These concept labels are typically provided by experts, which can be costly and require significant resources and effort. Additionally, concept saliency maps frequently misalign with input saliency maps, causing concept predictions to correspond to irrelevant input features - an issue related to annotation alignment. In this problem, we propose SSCBM, a strategy to generate pseudo labels and an alignment loss to solve these two problems. Results show our effectiveness.

{
    \small
    \bibliographystyle{plain}
    \bibliography{main}
}

\clearpage
\appendix
\section{Details of Experimental Setup}
\label{App:Setup}
\subsection{Datasets}\label{app:dataset}
We evaluate our methods on four real-world image tasks: \textit{CUB}, \textit{AwA2}, \textit{WBCatt} and \textit{7-point}.
\begin{itemize}
    \item \textbf{CUB \cite{he2019fine}}: the Caltech-UCSD Birds-200-2011 (CUB) dataset, which includes a total of 11,788 avian images including 4,796 training images 1,198 validation images and 5,794 testing images. The objective is to accurately categorize these birds into one of 200 distinct species. Following \cite{espinosa2022concept},  we use k = 112 binary bird attributes representing wing color, beak shape, etc.
    % \item \textbf{CelebA \cite{liu2015faceattributes}}: the Large-scale CelebFaces Attributes dataset, in the CelebA task, there are totally 202,599 face images from 6 balanced incomplete concept annotations and each image can be one of the 256 classes. We use 70\% images for training, 10\% images for validation and 20\% for testing.
    \item \textbf{AwA2 \cite{niu2018learning}}: Animals with Attributes 2 consists of in total 37,322 images distributed in 50 animal categories. We use 80\% for training, 10\% for validation and 10\% for testing. The AwA2 also provides a category-attribute matrix, which contains an 85-dim attribute vector (e.g., color, stripe, furry, size, and habitat) for each category.   
    \item  \textbf{WBCatt\cite{tsutsui2023wbcatt}}: the White Blood Cell Attributes dataset includes a total of 10,298 microscopic images from the PBC dataset \cite{pbc2019105020} with class label, include 6,169 images for training, 1,030 images for validation and 3,099 images for testing. Each image is annotated with 11 morphological attributes (e.g., cell shape, chromatin density and granule color). 
    \item\textbf{7-point\cite{Kawahara2018-7pt}}: the Seven-Point Checklist Dermatology Dataset is designed for diagnosing and classifying skin lesions. It includes 1011 lesion case distribute in 5 diagnostic categories. There are 413 samples for training, 203 samples for validation and 395 samples for testing. 
\end{itemize}
\begin{table}[htbp]
    \centering
    \resizebox{0.8\linewidth}{!}{
    \begin{tabular}{c|cccc}
        \toprule 
        %  & \textbf{CUB} & \textbf{CelebA} & \textbf{AwA2}&\textbf{WBCatt} &\textbf{7-point}\\
        % \midrule 
        % \text{Images} & 11,788 & 202,599 & 37,322 & 10,298 & 1011\\
        % \text{Classes} & 200 & 10,177 & 50 &5 & 5 \\
        % \text{Concepts} & 112 & 40 & 85 &31 & 19\\
        & \textbf{CUB} & \textbf{AwA2}&\textbf{WBCatt} &\textbf{7-point}\\
        \midrule 
        \text{Images} & 11,788 & 37,322 & 10,298 & 1011\\
        \text{Classes} & 200 & 50 & 5 & 5 \\
        \text{Concepts} & 112 & 85 &31 & 19\\

        \bottomrule
    \end{tabular}
}
    \caption{Statistics of the datasets used in our experiments.}
    \label{tab:datasets}
\end{table}

\subsection{Implementation Details}\label{app:imple_detail}
First, we resize the images to an input size of 299 x 299. Subsequently, we employ ResNet34 \cite{he2016deep} as the backbone to transform the input into latent code, followed by a fully connected layer to convert it into concept embeddings of size 16 (32 for CUB). During pseudo-labeling, we also utilize ResNet34 with the KNN algorithm with k = 2. Additionally, for obtaining concept labels using a threshold, we set the threshold to 0.6. We set $\lambda_1=1$ and $\lambda_2=0.1$ and utilize the SGD optimizer with a learning rate of 0.05 and a regularization coefficient of 5e-6. We train SSCBM for 100 epochs with a batch size of 256 (for AwA2, the batch size is 32 due to the large size of individual images). We repeat each experiment 5 times and report the average results. 

To construct the concept saliency map, we first upsample the heatmaps $\{\mathcal{H}_1, \mathcal{H}_2, \ldots, \mathcal{H}_k\}$ calculated in Section \ref{sec:unlabel} to the size $H \times W$ (the original image size). Then, we create a mask based on the value intensities, with higher values corresponding to darker colors.

% \subsection{}\label{app:saliency}
\subsection{Impact of Different Backbones}\label{app:backbone}
Here, we evaluate the performance of SSCBM and the baseline using different backbones (ResNet18, ResNet34, and ResNet50). We present the results in Table \ref{tab:backbone}. We observe that using ResNet34 as the backbone achieves a significant performance improvement compared to ResNet18. However, for ResNet50, its performance is almost on par with ResNet34, and in some cases, it even performs worse. We analyze that a possible reason could be that ResNet50 has a significantly larger number of parameters relative to the rest of the model, making it difficult to converge simultaneously with other parts during training, which leads to suboptimal results. We also note that \cite{espinosa2022concept} encounters a similar situation, where ResNet34 is used as the backbone.

\section{Test-time Intervention}
\label{app:testtime}
Test-time intervention enables human users to interact with the model in the inference time. We test our test-time intervention by correcting the 10\% to 100\% ratio of the concept label in the $\hat{g}$. For AwA2, there are no grouped concepts, so we adopt individual intervention. In the CUB, we do the group intervention, i.e., intervene in the concepts with associated attribution. For example, the breast color::yellow, breast color::black, and breast color::white are the same concept group. So, we only need to correct the concept label in the group. We expect that the model performance will steadily increase along with the ratio of concept intervention, indicating that the model learned such correct label information and automatically corrected other labels.  

% \begin{figure}[!htbp]
%     \centering
%     \includegraphics[width=\linewidth]{source/intervention.pdf}
%     \caption{Test-time Intervention on CUB and AwA2 dataset.
%     \label{fig:tti-celeba}}
% \end{figure}

\begin{figure}[htbp]
    \centering
    \begin{subfigure}{\linewidth}
        \centering
        \includegraphics[width=0.85\linewidth]{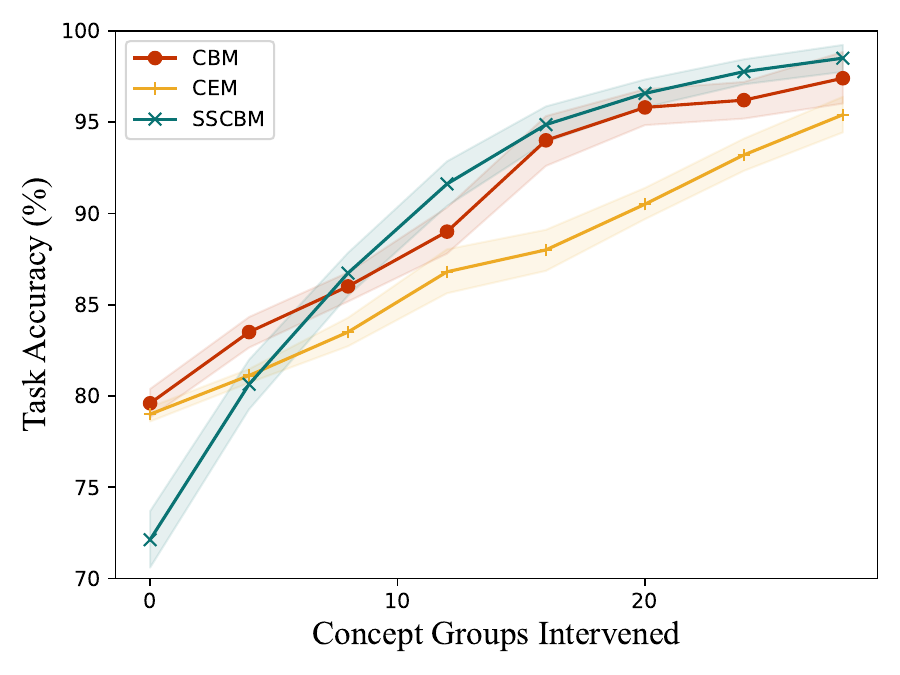}
        \caption{CUB}
        \label{fig:cub_int}
    \end{subfigure}
    
    \begin{subfigure}{\linewidth}
        \centering
        \includegraphics[width=0.85\linewidth]{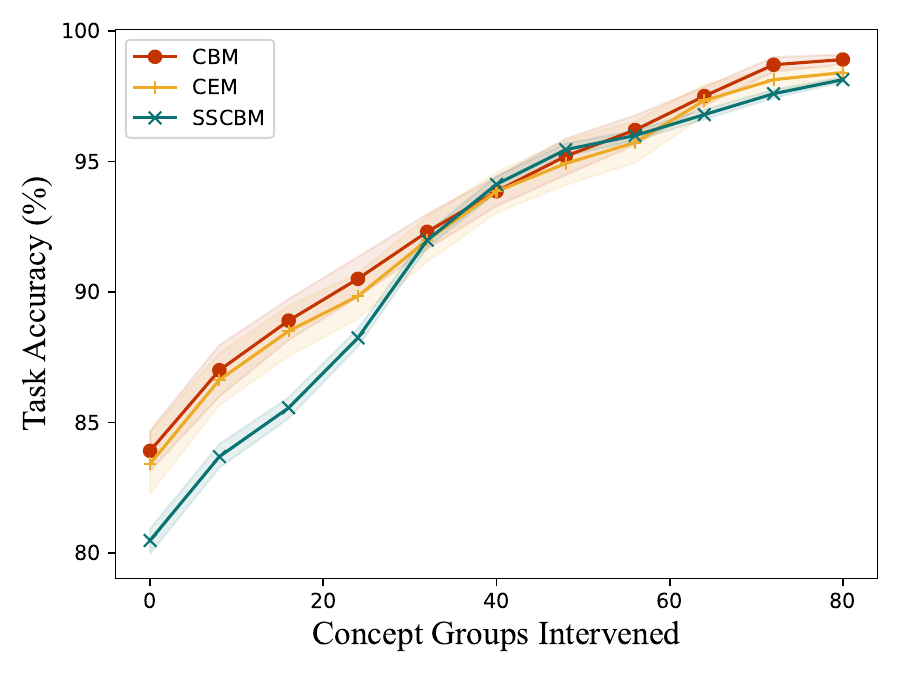}
        \caption{AwA2}
        \label{fig:AwA2_int}
    \end{subfigure}
    \caption{Test-time Intervention on CUB and AwA2 dataset.}
    \label{fig:intervention}
\end{figure}

Results in Figure \ref{fig:intervention} demonstrate our model's robustness and an increasing trend to learn the information of concept information, indicating our interpretability and model prediction performance. Here, we train SSCBM with a label ratio of 0.1 and compare its performance with CEM and CBM trained on the full dataset. It can be observed that without any intervention, the task performance of SSCBM is lower than the accuracy of the supervised model. However, as the number of intervened concept groups increases, the prediction accuracy of SSCBM gradually improves, eventually achieving comparable performance to CBM and CEM when interventions are applied to all concept groups. This lies in our loss of alignment in effectively learning the correct information pairs in unlabeled and labeled data.   

\begin{figure*}[t]
    \centering
    \includegraphics[width=0.8\linewidth]{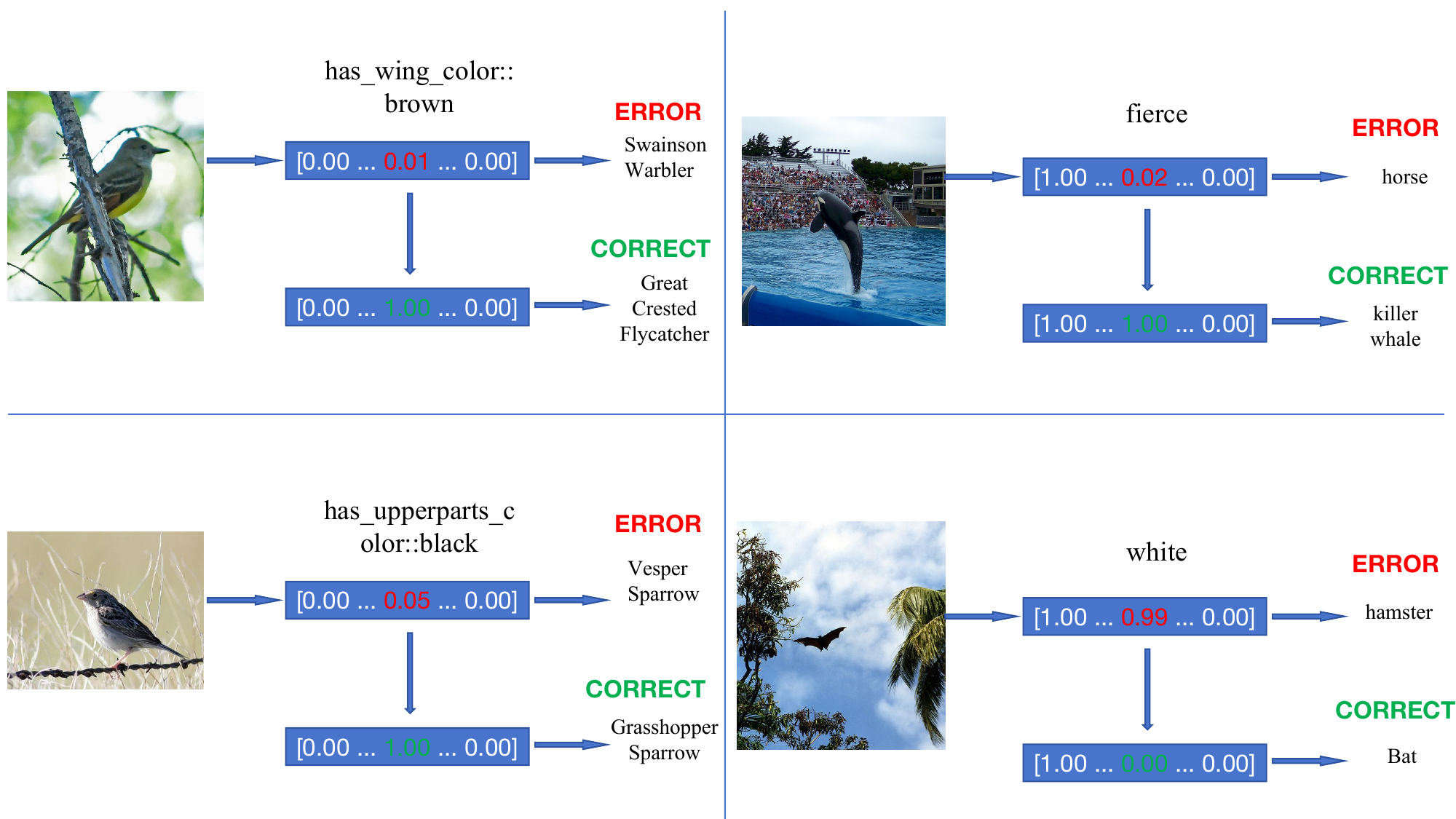}
    \caption{Examples of Test-time Intervention.}
    \label{tti-exp}
\end{figure*}

Here, we present some successful examples of Test-time Intervention illustrated in Figure \ref{tti-exp}. The first two on the left show examples from the CUB dataset. In the top left image, by changing the wing color to brown, we successfully caused the model to predict the Great Crested Flycatcher instead of the Swainson Warbler. In the bottom left, because the model initially failed to notice that the upper part of the bird was black, it misclassified it as Vesper Sparrow. Through test-time intervention, we successfully made it predicted the bird was a Grasshopper Sparrow. The results on the right side of the image are from the AwA2 dataset. We successfully made the model predict correctly by modifying concepts at test time. For example, in the top right image, by modifying the concept of 'fierce' for the orca, we prevented it from being predicted as a horse. In the bottom right, we successfully made the model recognize the bat through the color of the bat.

\begin{table*}[htbp]
\centering
\caption{Performance of different backbones under different ratios of labeled data.}\label{tab:backbone}

\resizebox{0.65\linewidth}{!}{
\begin{tabular}{ccccccccc}
\toprule
\multirow{2}{*}{\textbf{Dataset}} & \multirow{2}{*}{\textbf{Backbone}} & \multirow{2}{*}{\textbf{Ratio}} & \multicolumn{2}{c}{\textbf{CBM+SSL}} & \multicolumn{2}{c}{\textbf{CEM+SSL}} &  \multicolumn{2}{c}{\textbf{SSCBM}} \\
\cmidrule(lr){4-5} \cmidrule(lr){6-7} \cmidrule(lr){8-9} 
& & & \textbf{Concept} & \textbf{Task} & \textbf{Concept} & \textbf{Task} & \textbf{Concept} & \textbf{Task} \\
\midrule

%%%%%%%%%%%%%%%%%%%%%%%%%%%%%%%%%%%%%%%%%%%%%%%%%%%%%%%%%%
\multirow{15}{*}{CUB} &
\multirow{5}{*}{ResNet18} 
&K=1 & 81.69\% & 7.92\%  & 80.44\% & 54.90\% & 87.50\% & 60.77\% \\
&&0.05 (K=2)     & 84.67\% & 7.46\%  & 83.16\% & 60.53\% & 89.16\% & 64.31\% \\
&&0.1 (K=3)      & 85.01\% & 8.54\%  & 82.46\% & 59.56\% & 90.28\% & 66.76\% \\
&&0.15 (K=4)    & 85.17\% & 10.20\% & 83.81\% & 64.27\% & 90.55\% & 66.40\% \\
&&0.2 (K=5)     & 85.42\% & 9.70\%  & 84.35\% & 64.22\% & 91.02\% & 68.31\%\\

\cmidrule(lr){2-9}

& \multirow{5}{*}{ResNet34} 
 & K=1 & 83.11\% & 5.51\% & 82.36\% & 59.35\%  & 88.99\% & 66.72\% \\
& &0.05 (K=2) & 84.51\% & 8.35\% & 83.72\% & 62.20\%  & 90.04\% & 67.43\% \\
& &0.1 (K=3) & 84.96\% & 9.84\% & 84.03\% & 63.12\% & 90.88\% & 67.67\% \\
& &0.15 (K=4) & 85.47\% & 9.96\% & 84.30\% & 64.14\%  & 91.47\% & 68.36\% \\
& &0.2 (K=5) & 86.67\% & 16.43\%& 86.83\% &67.64\% & 92.09 \% & 70.07\% \\

\cmidrule(lr){2-9}
& \multirow{5}{*}{ResNet50} 
&K=1 & 81.43\% & 9.75\%  & 78.99\% & 57.39\% & 89.03\% & 69.59\% \\
&&0.05 (K=2)     & 84.49\% & 8.42\%  & 83.00\% & 62.12\% & 90.91\% & 71.73\% \\
&&0.1 (K=3)      & 85.09\% & 7.97\%  & 83.15\% & 63.27\% & 92.10\% & 74.09\% \\
&&0.15 (K=4)    & 85.14\% & 9.56\%  & 83.32\% & 64.43\% & 91.75\% & 68.50\% \\
&&0.2 (K=5)     & 85.39\% & 10.67\% & 83.71\% & 66.09\% & 93.21\% & 75.77\%\\

%%%%%%%%%%%%%%%%%%%%%%%%%%%%%%%%%%%%%%%%%%%%%%%%%%%%%%%%%%
\cmidrule(lr){1-9}
\multirow{15}{*}{WBCatt} &
\multirow{5}{*}{ResNet18} 
&K=1 & 72.21\% & 99.45\% & 73.06\% & 99.42\% & 89.07\% & 99.58\% \\
&&0.05 (K=62)     & 80.83\% & 99.65\% & 76.75\% & 99.19\% & 93.76\% & 99.55\% \\
&&0.1 (K=124)      & 83.33\% & 99.81\% & 78.74\% & 98.87\% & 93.54\% & 99.55\% \\
&&0.15 (K=186)    & 85.10\% & 99.84\% & 82.96\% & 99.52\% & 93.83\% & 99.55\% \\
&&0.2 (K=247)     & 84.87\% & 99.74\% & 84.11\% & 99.42\% & 94.18\% & 99.58\%\\

\cmidrule(lr){2-9}

& \multirow{5}{*}{ResNet34} 
&K=1 & 79.06\% & 99.39\% & 70.27\% & 98.64\% & 91.48\% & 99.13\% \\
&&0.05 (K=62)     & 81.08\% & 99.48\% & 73.82\% & 99.52\% & 93.53\% & 99.61\% \\
&&0.1 (K=124)      & 85.48\% & 99.32\% & 72.25\% & 99.29\% & 93.98\% & 99.68\% \\
&&0.15 (K=186)    & 85.39\% & 99.68\% & 72.68\% & 99.58\% & 94.42\% & 99.71\% \\
&&0.2 (K=247)     & 87.07\% & 99.74\% & 74.14\% & 99.52\% & 94.42\% & 99.71\%\\

\cmidrule(lr){2-9}
& \multirow{5}{*}{ResNet50} 
&K=1 & 71.30\% & 99.71\% & 72.99\% & 99.48\% & 88.46\% & 99.71\% \\
&&0.05 (K=62)     & 82.69\% & 99.45\% & 69.74\% & 99.35\% & 93.73\% & 99.61\% \\
&&0.1 (K=124)      & 81.23\% & 99.71\% & 72.22\% & 98.32\% & 93.99\% & 99.55\% \\
&&0.15 (K=186)    & 82.97\% & 99.74\% & 87.17\% & 99.23\% & 94.32\% & 99.71\% \\
&&0.2 (K=247)     & 84.40\% & 99.84\% & 85.03\% & 99.77\% & 94.47\% & 99.61\% \\
%%%%%%%%%%%%%%%%%%%%%%%%%%%%%%%%%%%%%%%%
\cmidrule(lr){1-9}

\multirow{15}{*}{7-point} &
\multirow{5}{*}{ResNet18} 
& K=1 & 56.08\% & 56.46\% & 56.30\% & 61.52\% & 58.19\% & 64.56\% \\
& & 0.05 (K=5)   & 63.58\% & 59.75\% & 64.32\% & 62.28\% & 67.24\% & 66.08\% \\
& & 0.1 (K=9)    & 65.78\% & 63.29\% & 66.65\% & 64.30\% & 71.03\% & 68.61\% \\
& & 0.15 (K=13)  & 67.71\% & 63.54\% & 67.02\% & 67.34\% & 73.34\% & 69.87\% \\
& & 0.2 (K=17)   & 69.54\% & 57.22\% & 68.79\% & 63.80\% & 75.02\% & 70.63\% \\

\cmidrule(lr){2-9}

& \multirow{5}{*}{ResNet34} 
& K=1 & 59.91\% & 55.95\% & 62.78\% & 66.09\% & 66.58\% & 66.84\% \\
& & 0.05 (K=5)   & 65.36\% & 57.47\% & 67.85\% & 67.09\% & 70.98\% & 68.77\% \\
& & 0.1 (K=9)    & 68.82\% & 55.70\% & 72.23\% & 66.33\% & 73.67\% & 70.09\% \\
& & 0.15 (K=13)  & 66.14\% & 59.75\% & 66.54\% & 67.09\% & 73.94\% & 72.56\% \\
& & 0.2 (K=17)   & 70.29\% & 55.44\% & 73.04\% & 66.84\% & 76.52\% & 74.56\% \\

\cmidrule(lr){2-9}
& \multirow{5}{*}{ResNet50} 
&K=1 & 55.95\% & 64.81\% & 56.46\% & 64.56\% & 59.89\% & 64.30\% \\
&&0.05 (K=5)     & 62.80\% & 60.00\% & 64.66\% & 66.08\% & 69.29\% & 66.84\% \\
&&0.1 (K=9)      & 65.14\% & 60.25\% & 66.48\% & 67.34\% & 68.54\% & 64.56\% \\
&&0.15 (K=13)    & 69.15\% & 57.47\% & 67.94\% & 66.33\% & 73.58\% & 67.85\% \\
&&0.2 (K=17)     & 71.03\% & 58.73\% & 67.97\% & 67.09\% & 74.27\% & 64.05\%\\

\bottomrule

\end{tabular}}

\end{table*}

\begin{figure*}[htbp] 
    \setlength{\tabcolsep}{0.2pt} % Default value: 6pt
    \renewcommand{\arraystretch}{1.0} % Default value: 1
    \begin{center}
    \resizebox{0.8\linewidth}{!}{
    \begin{tabular*}{\linewidth}
    {@{\extracolsep{\fill}}cccc}
    \includegraphics[width=0.249\linewidth]{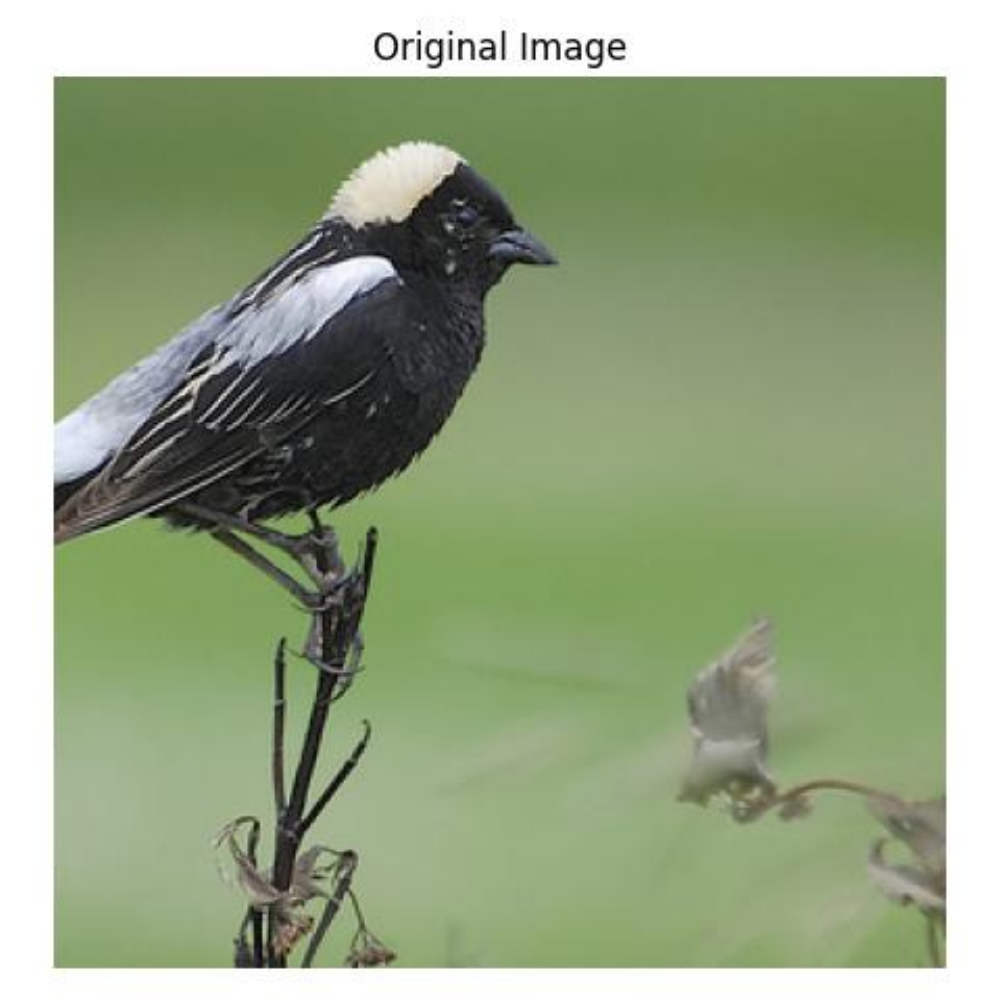} &
    \includegraphics[width=0.249\linewidth]{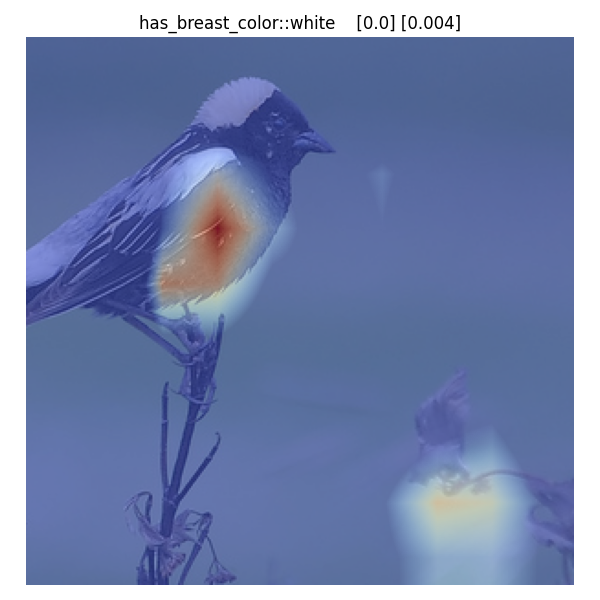} &
    \includegraphics[width=0.249\linewidth]{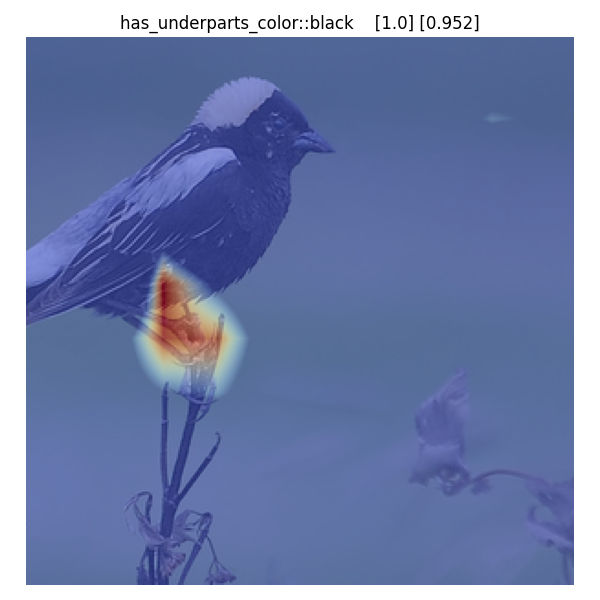} &
    \includegraphics[width=0.249\linewidth]{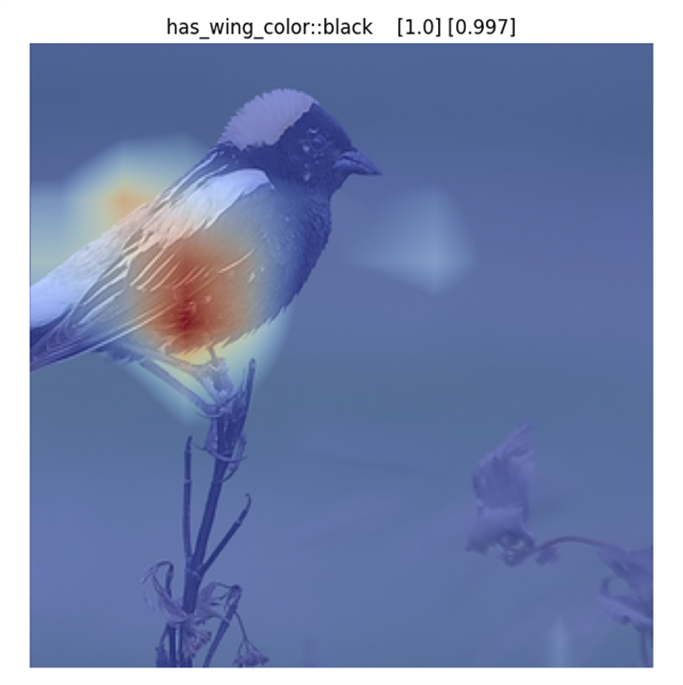} \\  
    \end{tabular*}}
    \vspace{-12pt}
    \caption{Concept saliency map on CUB dataset (bobolink) shows reasonable localization of the ground truth concept regions in the input image. \label{fig:bird4}}
    \end{center}
\vspace{-12pt}
\end{figure*} 
\begin{figure*}[htbp] 
    \setlength{\tabcolsep}{0.2pt} % Default value: 6pt
    \renewcommand{\arraystretch}{1.0} % Default value: 1
    \begin{center}
    \resizebox{0.8\linewidth}{!}{
    \begin{tabular*}{\linewidth}
    {@{\extracolsep{\fill}}cccc}
    \includegraphics[width=0.249\linewidth]{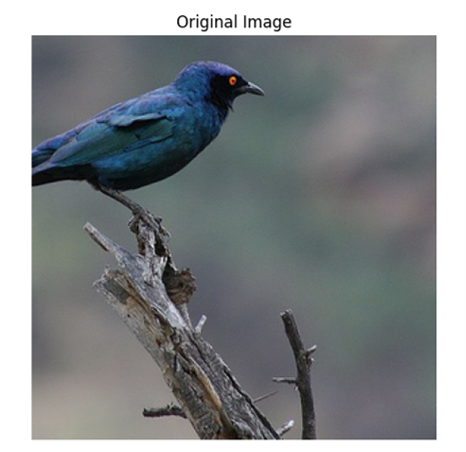} &
    \includegraphics[width=0.249\linewidth]{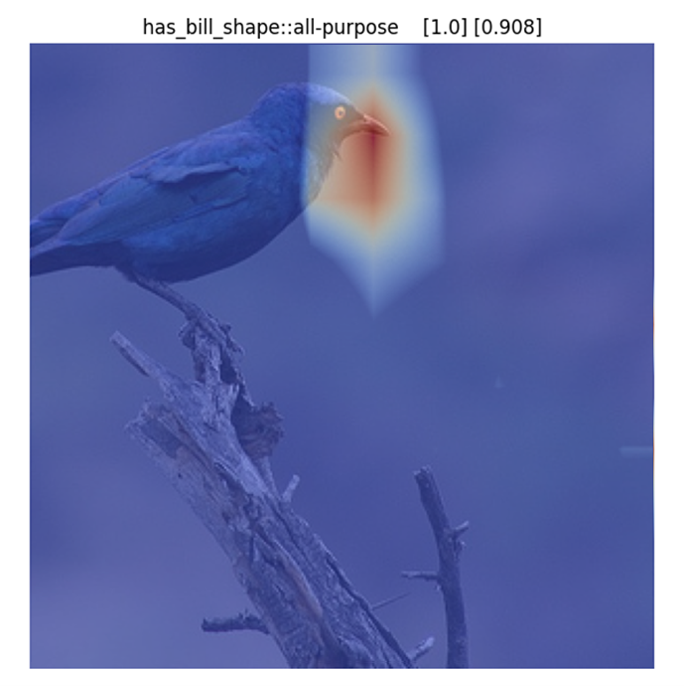} &
    \includegraphics[width=0.249\linewidth]{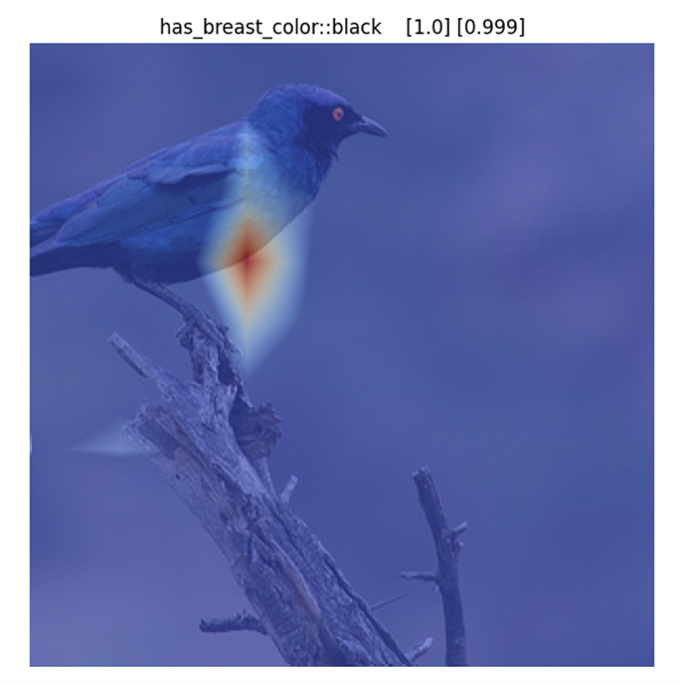} &
    \includegraphics[width=0.249\linewidth]{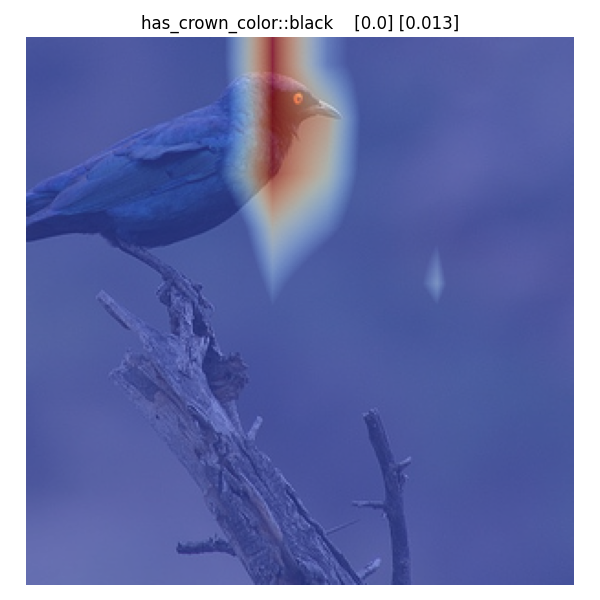} \\  
    \end{tabular*}}
    \vspace{-12pt}
    \caption{Concept saliency map on CUB dataset (cape glossy starling) shows reasonable localization of the ground truth concept regions in the input image. \label{fig:bird5}}
    \end{center}
\vspace{-12pt}
\end{figure*} 
\begin{figure*}[htbp] 
    \setlength{\tabcolsep}{0.2pt} % Default value: 6pt
    \renewcommand{\arraystretch}{1.0} % Default value: 1
    \begin{center}
    \resizebox{0.8\linewidth}{!}{
    \begin{tabular*}{\linewidth}
    {@{\extracolsep{\fill}}cccc}
    \includegraphics[width=0.249\linewidth]{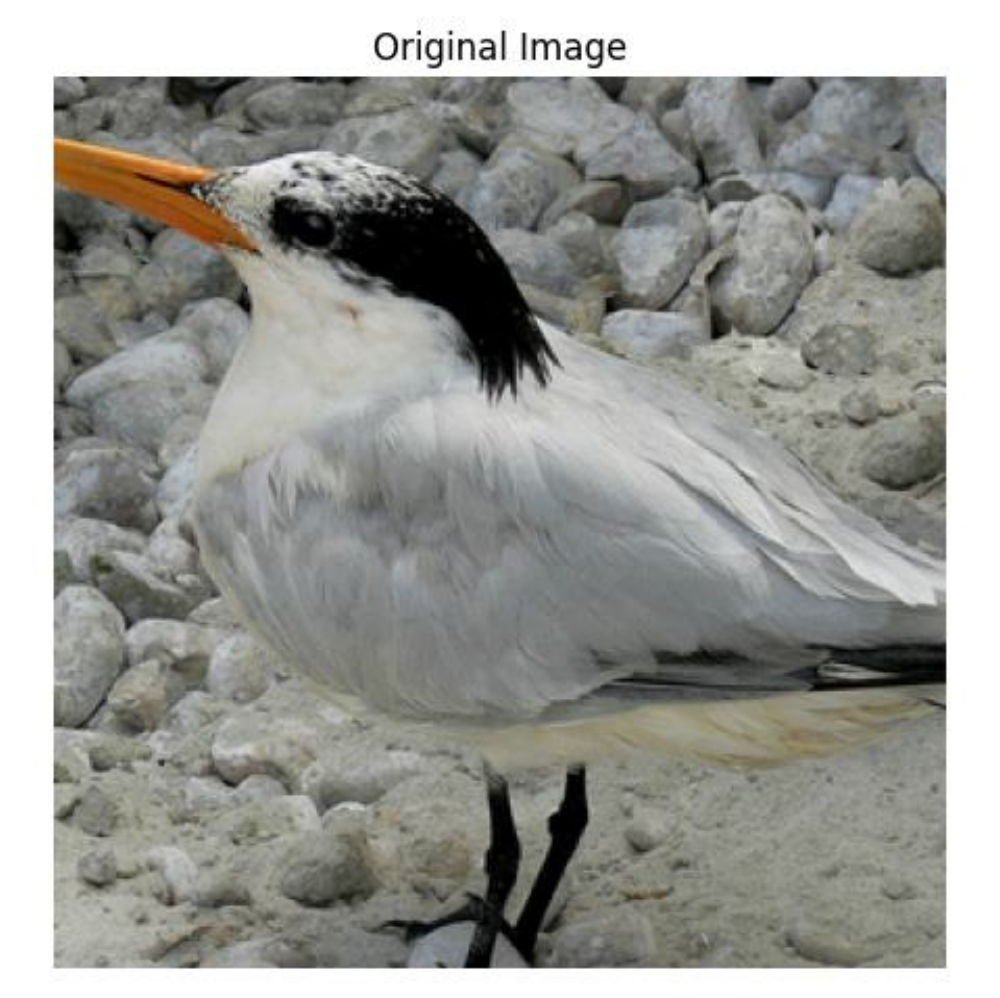} &
    \includegraphics[width=0.249\linewidth]{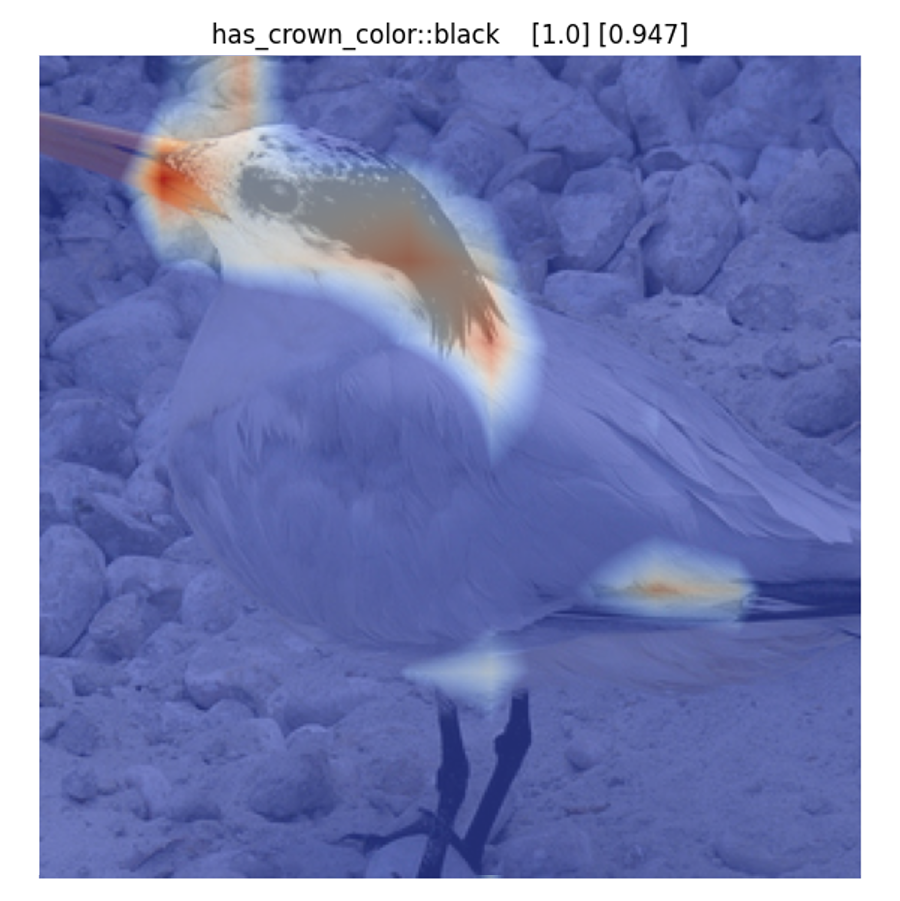} &
    \includegraphics[width=0.249\linewidth]{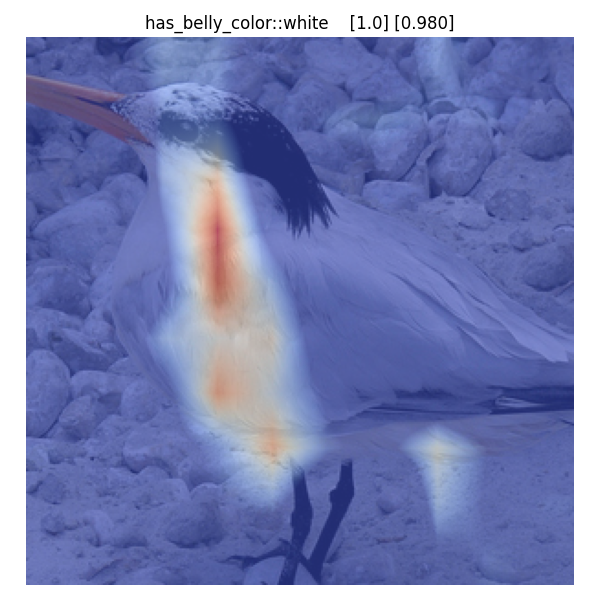} &
    \includegraphics[width=0.249\linewidth]{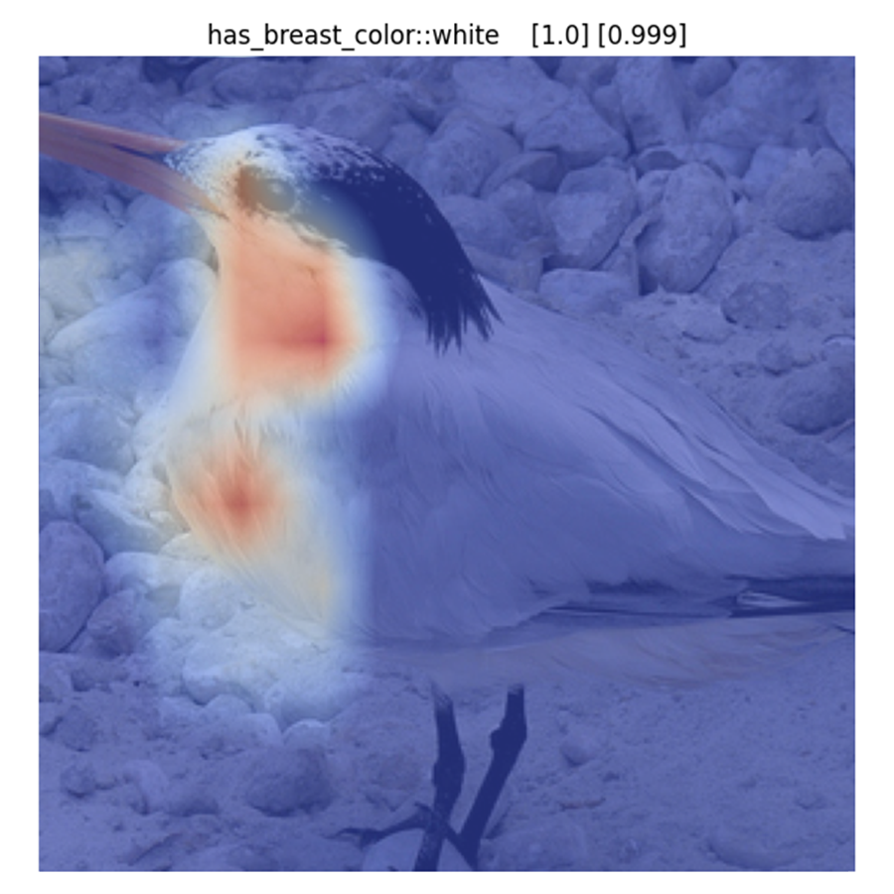} \\  
    \end{tabular*}}
    \vspace{-12pt}
    \caption{Concept saliency map on CUB dataset (elegant tern) shows reasonable localization of the ground truth concept regions in the input image. \label{fig:bird6}}
    \end{center}
\vspace{-12pt}
\end{figure*} 
\begin{figure*}[htbp] 
    \setlength{\tabcolsep}{0.2pt} % Default value: 6pt
    \renewcommand{\arraystretch}{1.0} % Default value: 1
    \begin{center}
    \resizebox{0.8\linewidth}{!}{
    \begin{tabular*}{\linewidth}
    {@{\extracolsep{\fill}}cccc}
    \includegraphics[width=0.249\linewidth]{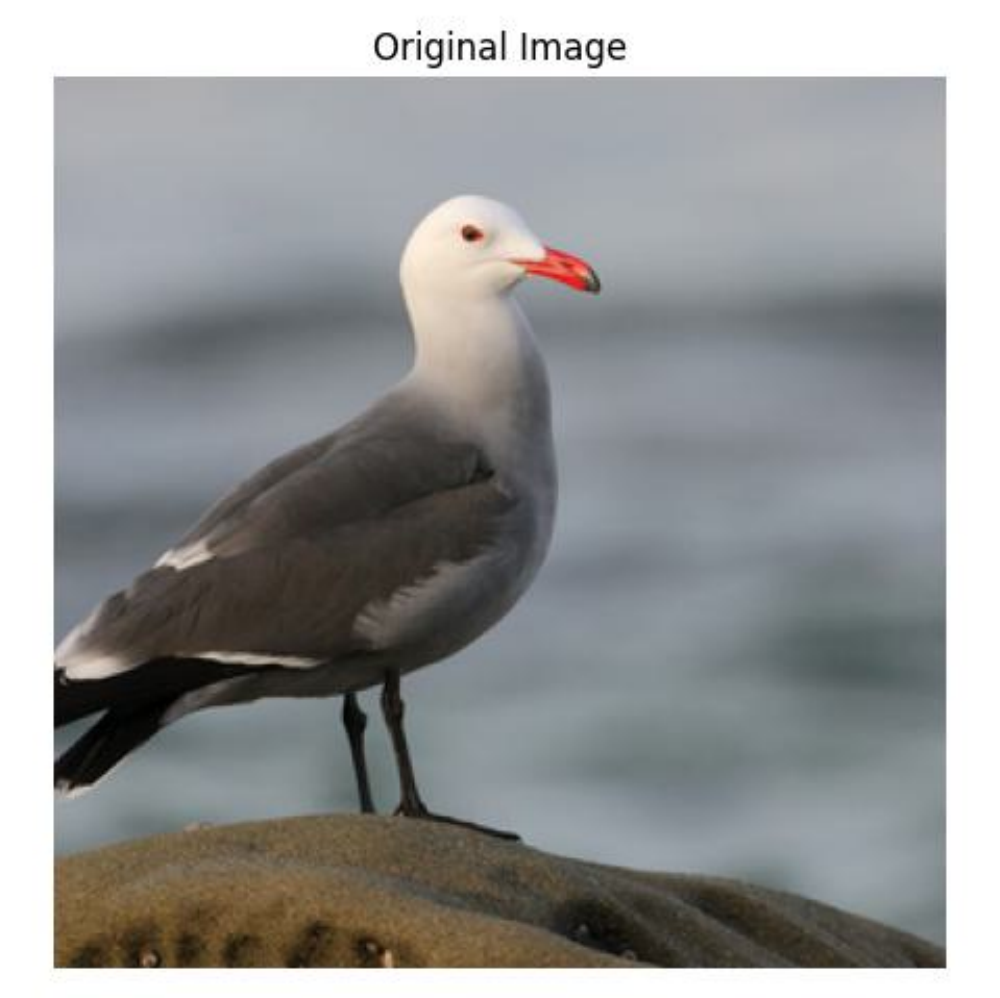} &
    \includegraphics[width=0.249\linewidth]{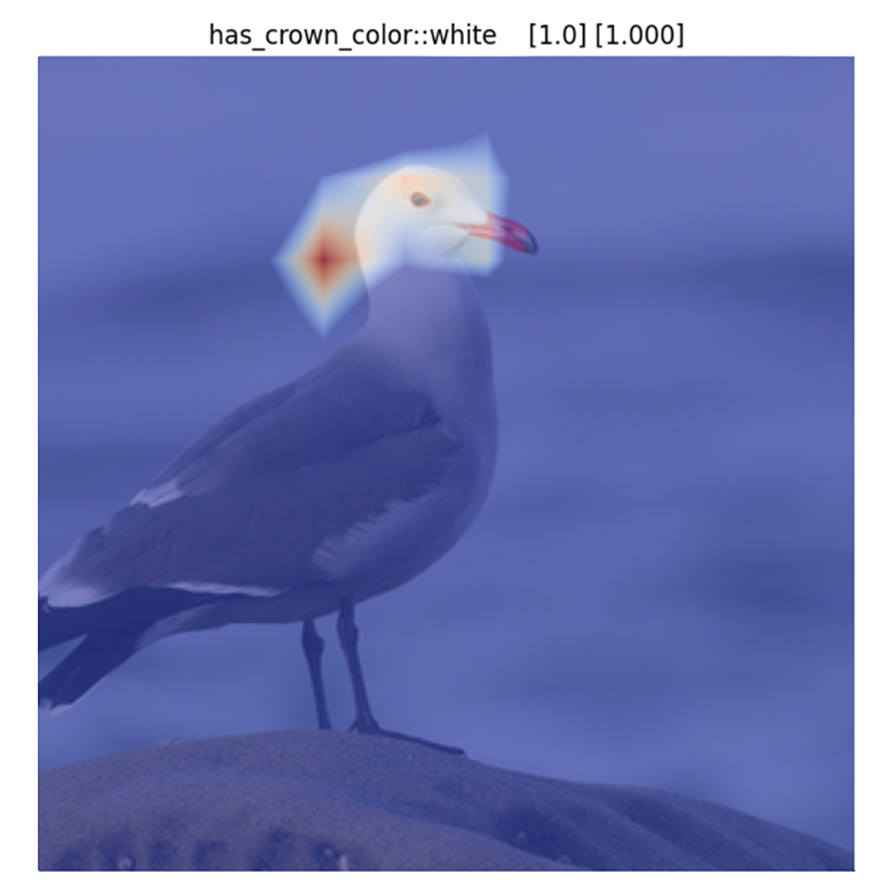} &
    \includegraphics[width=0.249\linewidth]{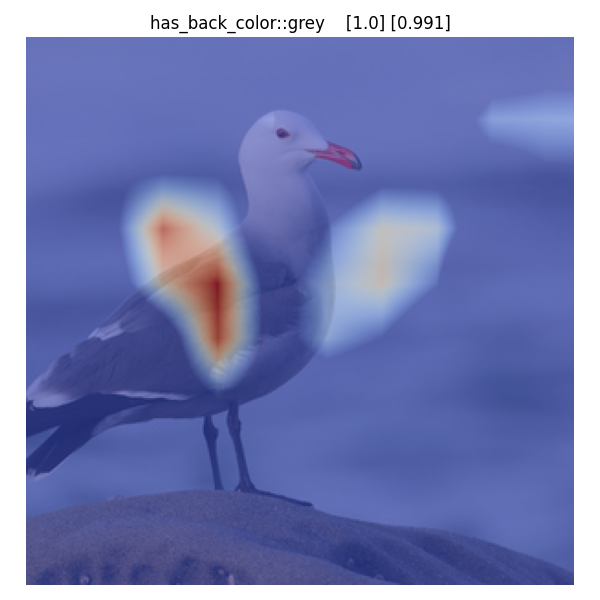} &
    \includegraphics[width=0.249\linewidth]{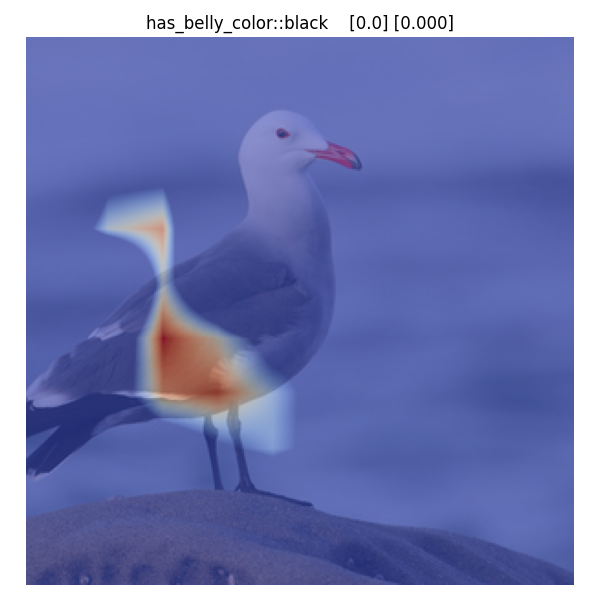} \\  
    \end{tabular*}}
    \vspace{-12pt}
    \caption{Concept saliency map on CUB dataset (heermann gull) shows reasonable localization of the ground truth concept regions in the input image. \label{fig:bird7}}
    \end{center}
\vspace{-12pt}
\end{figure*} 
\begin{figure*}[htbp] 
    \setlength{\tabcolsep}{0.2pt} % Default value: 6pt
    \renewcommand{\arraystretch}{1.0} % Default value: 1
    \begin{center}
    \resizebox{0.8\linewidth}{!}{
    \begin{tabular*}{\linewidth}
    {@{\extracolsep{\fill}}cccc}
    \includegraphics[width=0.249\linewidth]{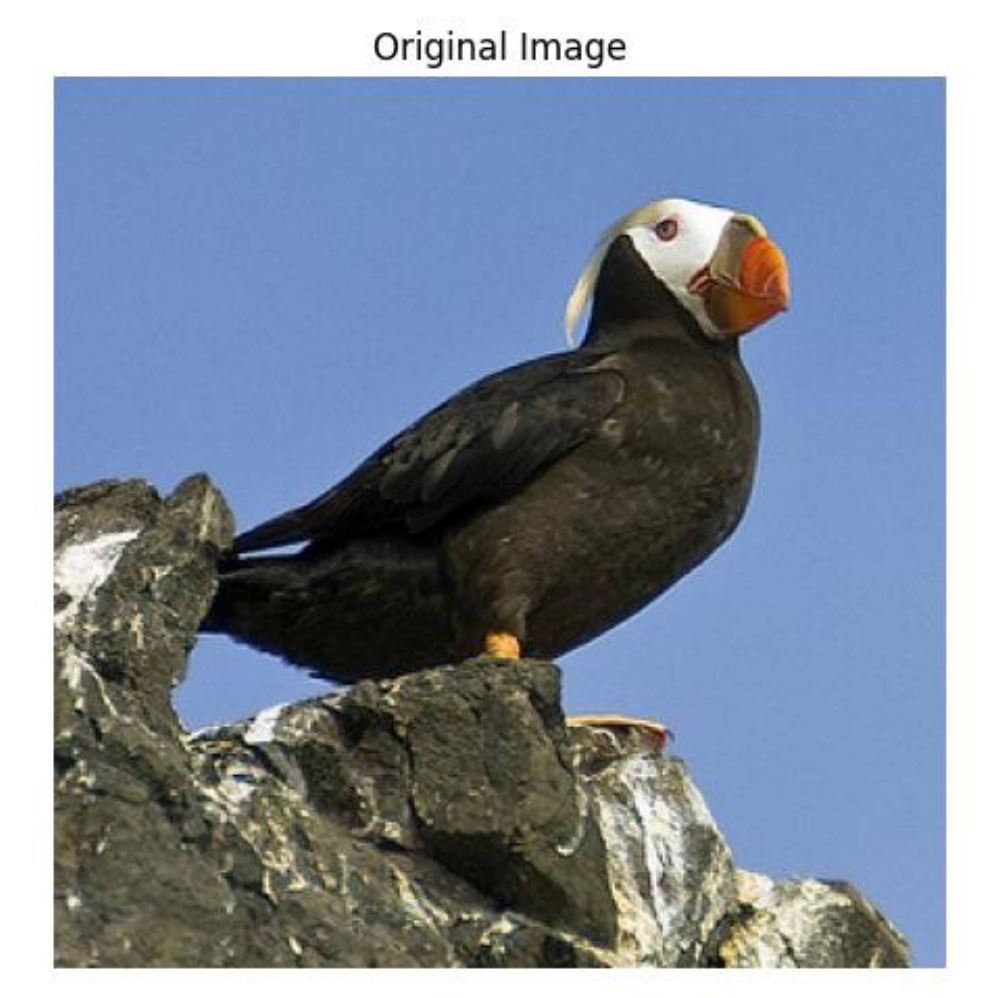} &
    \includegraphics[width=0.249\linewidth]{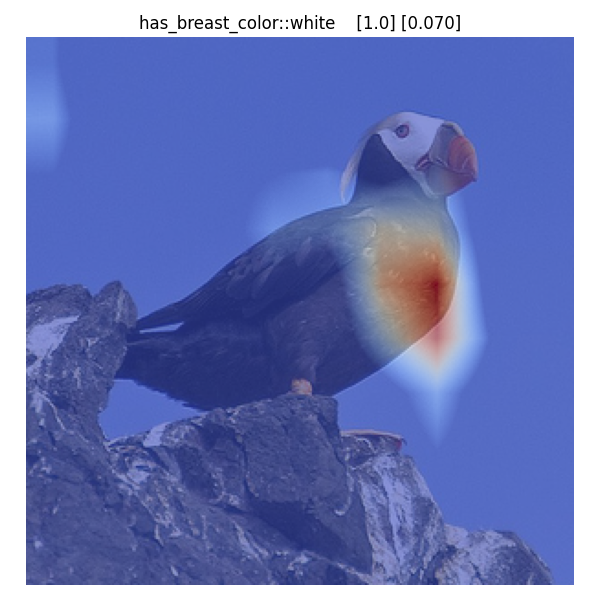} &
    \includegraphics[width=0.249\linewidth]{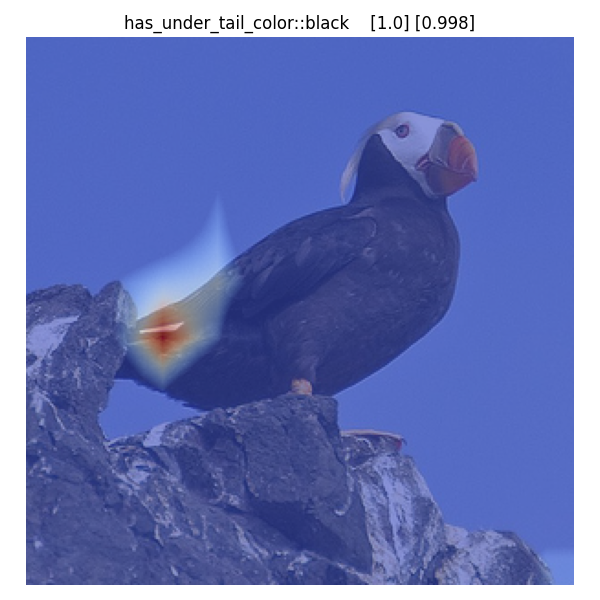} &
    \includegraphics[width=0.249\linewidth]{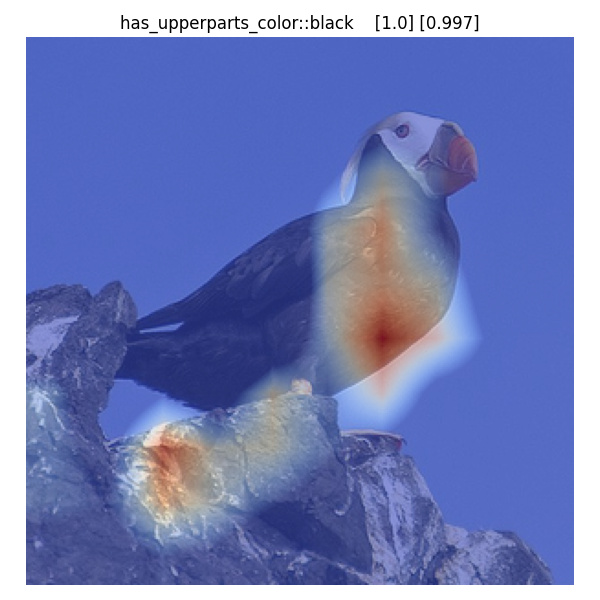} \\  
    \end{tabular*}}
    \vspace{-12pt}
    \caption{Concept saliency map on CUB dataset (horned puffin) shows reasonable localization of the ground truth concept regions in the input image. \label{fig:bird8}}
    \end{center}
\vspace{-12pt}
\end{figure*} 
\begin{figure*}[htbp] 
    \setlength{\tabcolsep}{0.2pt} % Default value: 6pt
    \renewcommand{\arraystretch}{1.0} % Default value: 1
    \begin{center}
    \resizebox{0.8\linewidth}{!}{
    \begin{tabular*}{\linewidth}
    {@{\extracolsep{\fill}}cccc}
    \includegraphics[width=0.249\linewidth]{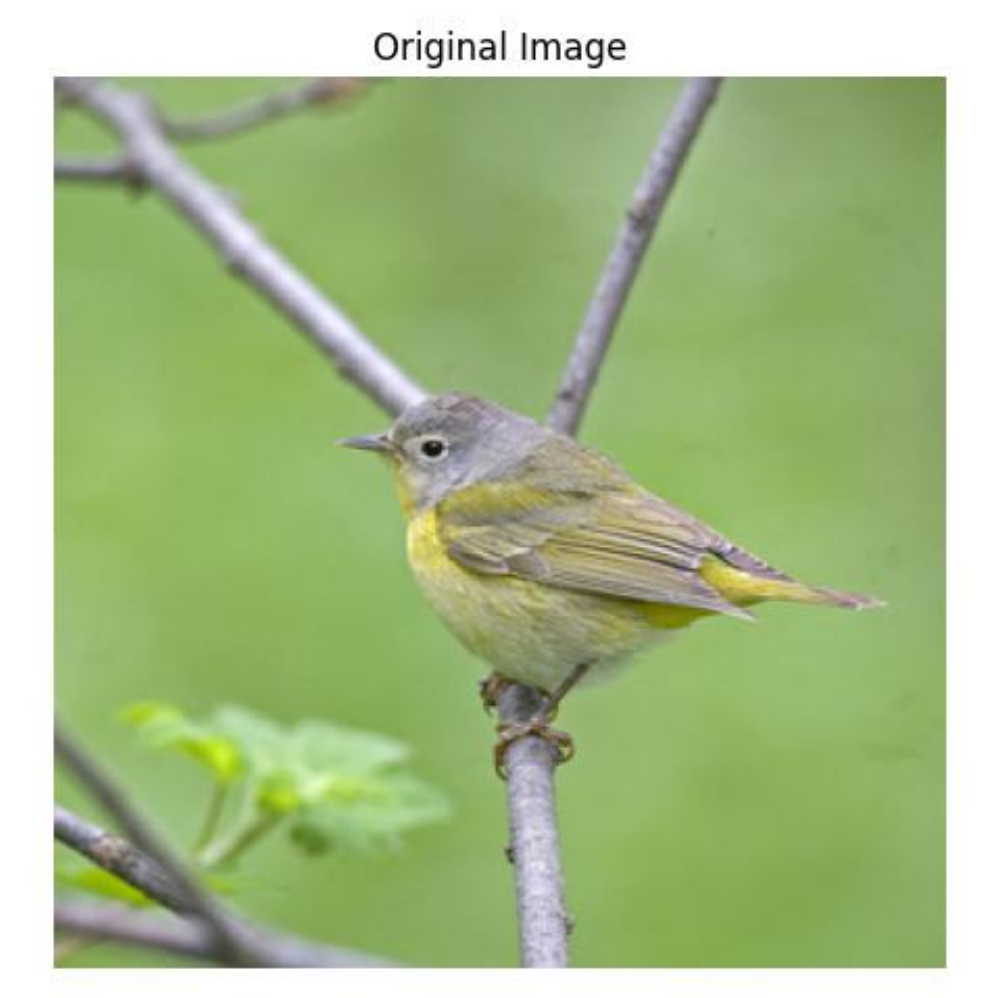} &
    \includegraphics[width=0.249\linewidth]{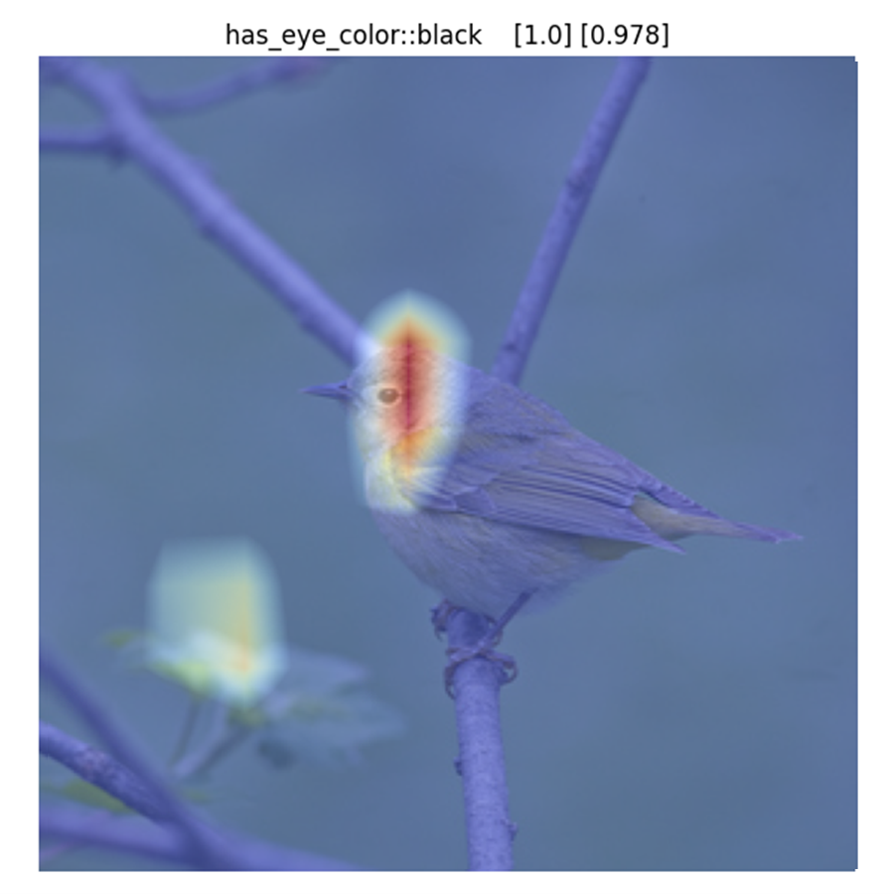} &
    \includegraphics[width=0.249\linewidth]{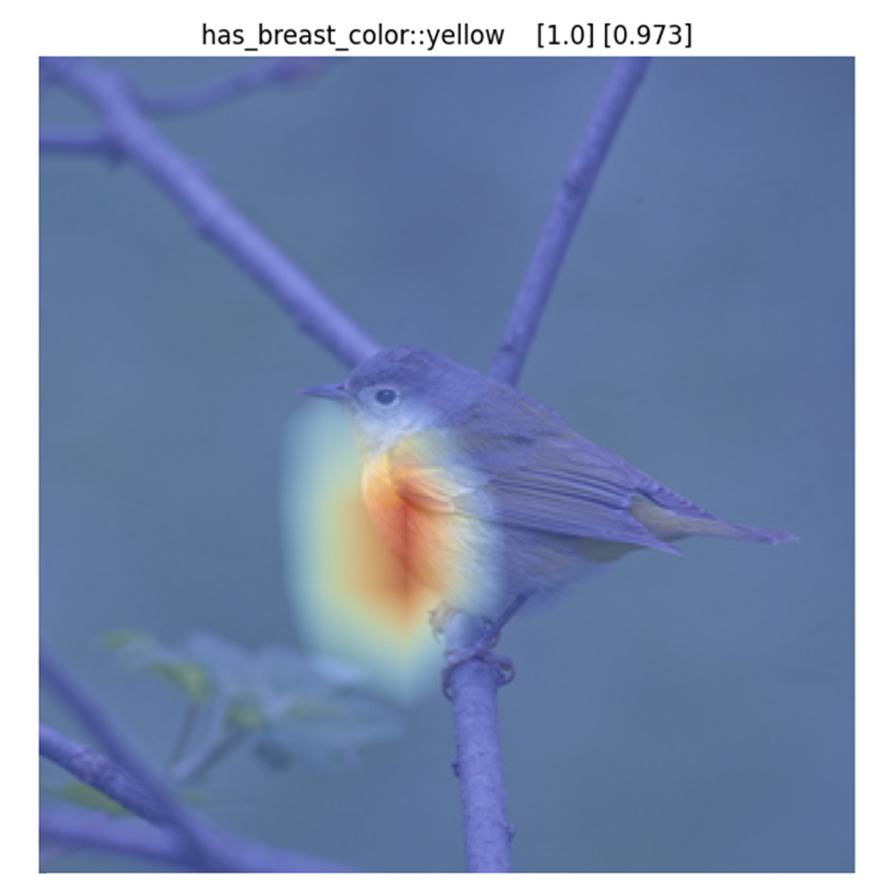} &
    \includegraphics[width=0.249\linewidth]{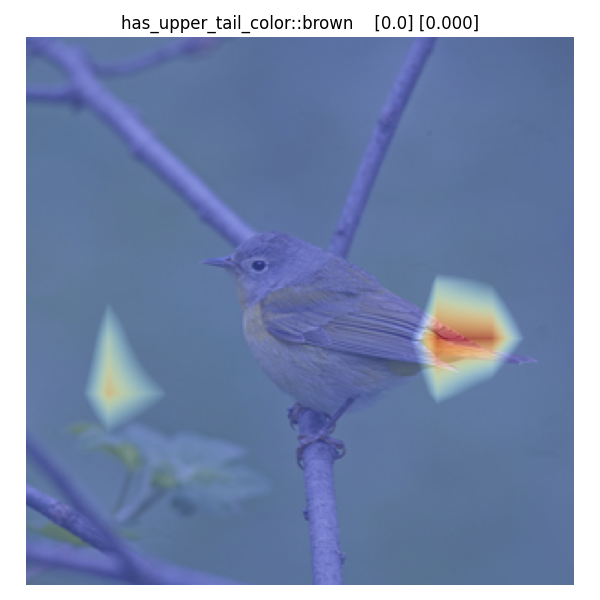} \\  
    \end{tabular*}}
    \vspace{-12pt}
    \caption{Concept saliency map on CUB dataset (nashville warbler) shows reasonable localization of the ground truth concept regions in the input image. \label{fig:bird9}}
    \end{center}
\vspace{-12pt}
\end{figure*} 
\begin{figure*}[htbp] 
    \setlength{\tabcolsep}{0.2pt} % Default value: 6pt
    \renewcommand{\arraystretch}{1.0} % Default value: 1
    \begin{center}
    \resizebox{0.8\linewidth}{!}{
    \begin{tabular*}{\linewidth}
    {@{\extracolsep{\fill}}cccc}
    \includegraphics[width=0.249\linewidth]{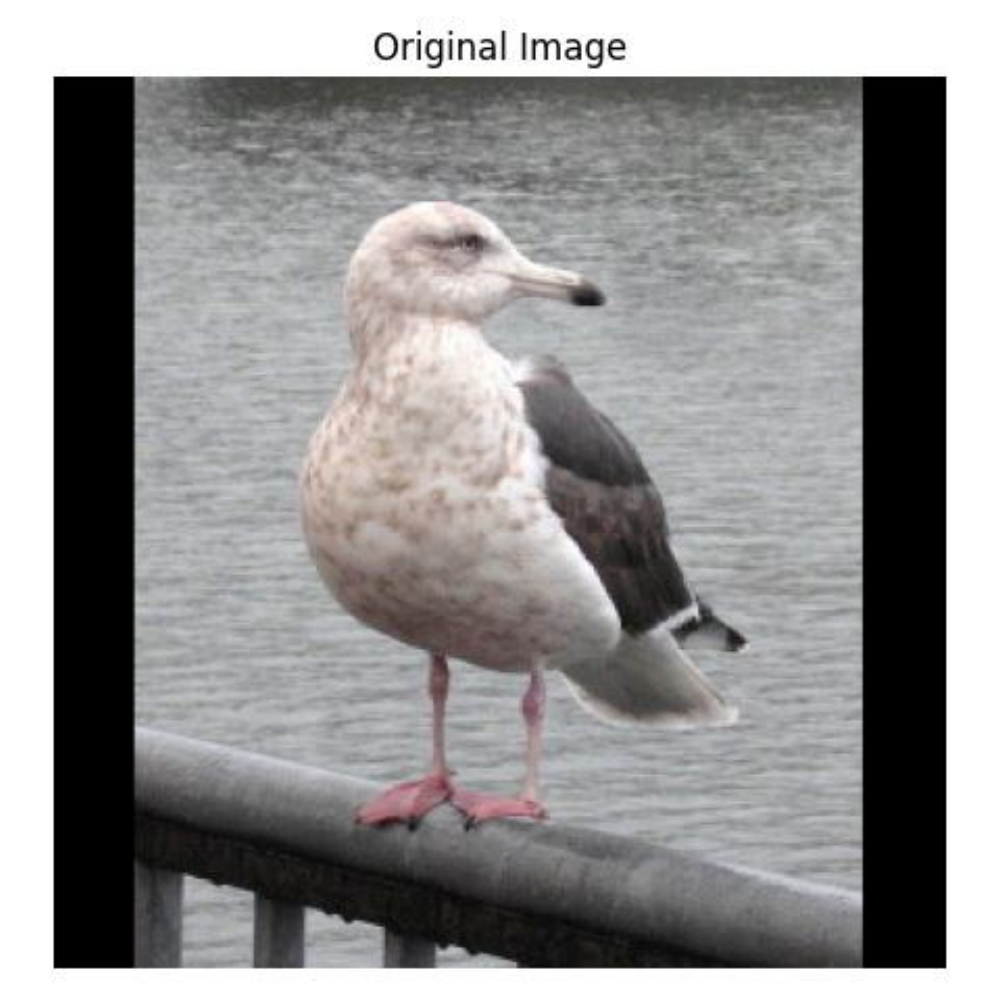} &
    \includegraphics[width=0.249\linewidth]{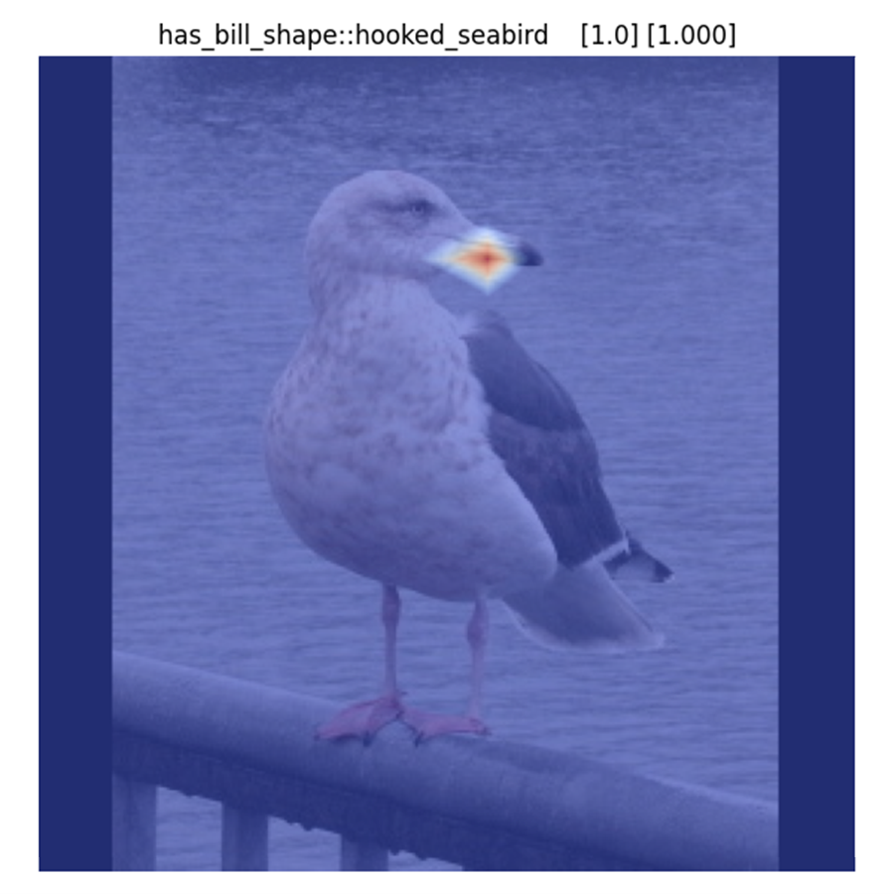} &
    \includegraphics[width=0.249\linewidth]{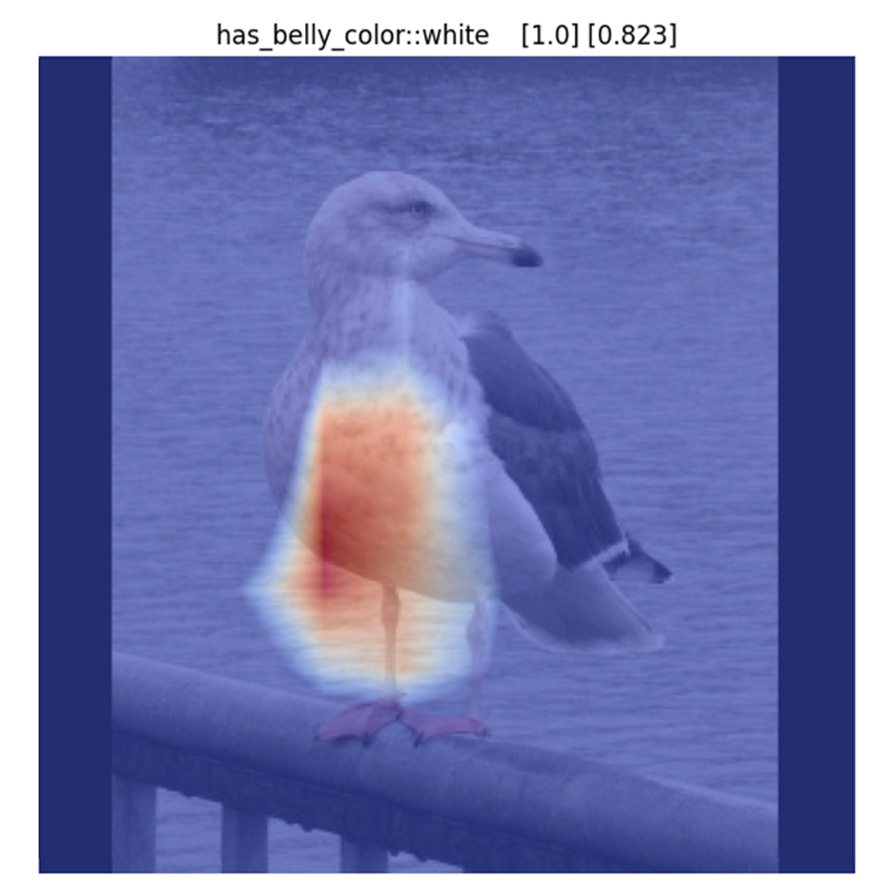} &
    \includegraphics[width=0.249\linewidth]{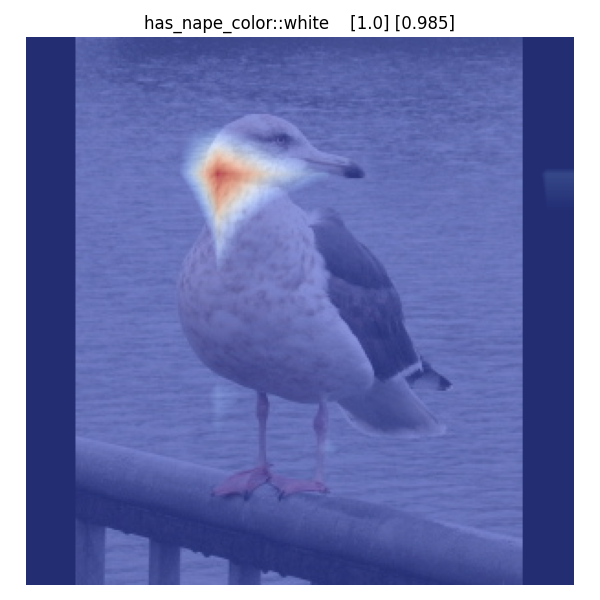} \\  
    \end{tabular*}}
    \vspace{-12pt}
    \caption{Concept saliency map on CUB dataset (slaty backed gull) shows reasonable localization of the ground truth concept regions in the input image. \label{fig:bird10}}
    \end{center}
\vspace{-12pt}
\end{figure*} 
\begin{figure*}[htbp] 
    \setlength{\tabcolsep}{0.2pt} % Default value: 6pt
    \renewcommand{\arraystretch}{1.0} % Default value: 1
    \begin{center}
    \resizebox{0.8\linewidth}{!}{
    \begin{tabular*}{\linewidth}
    {@{\extracolsep{\fill}}cccc}
    \includegraphics[width=0.249\linewidth]{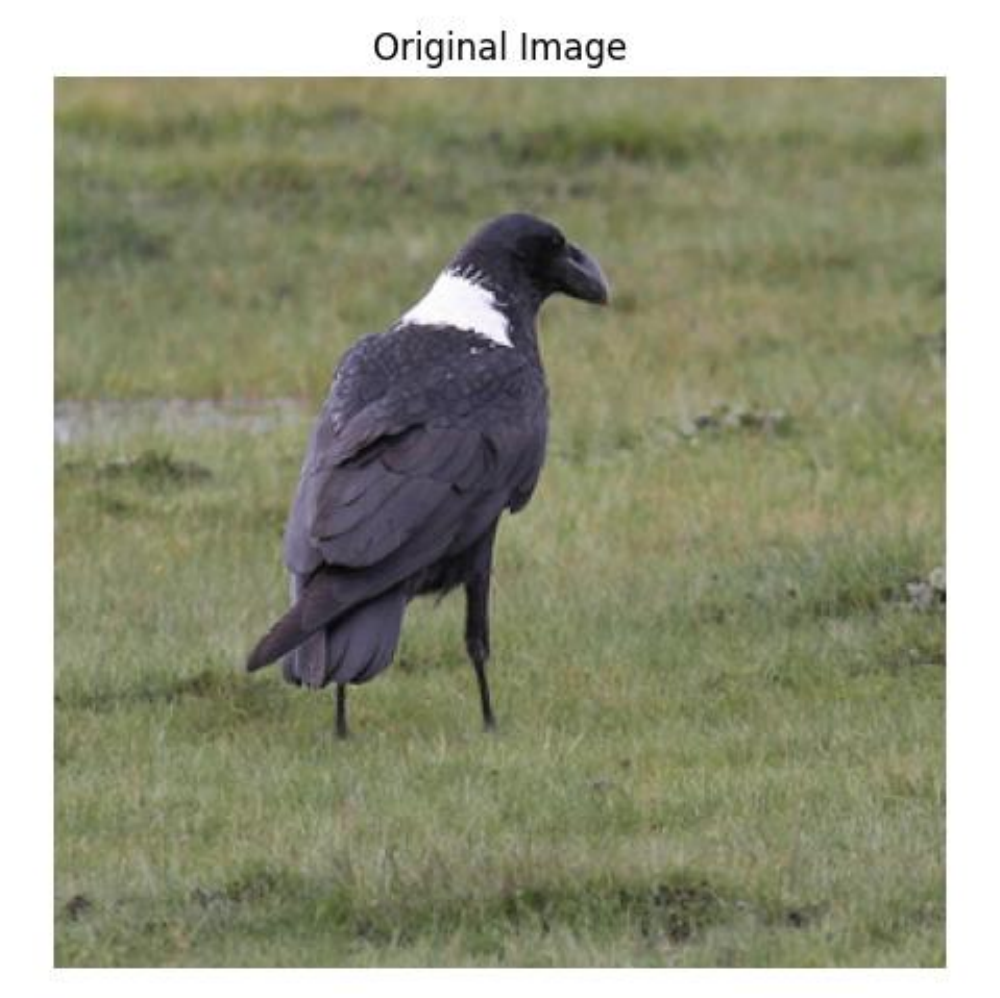} &
    \includegraphics[width=0.249\linewidth]{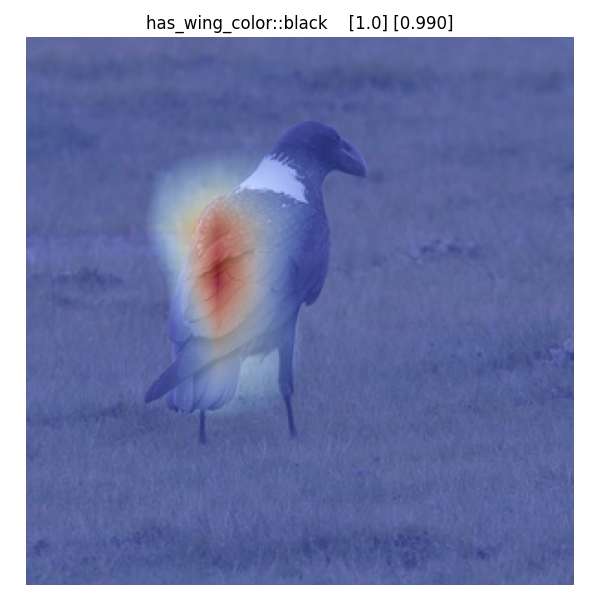} &
    \includegraphics[width=0.249\linewidth]{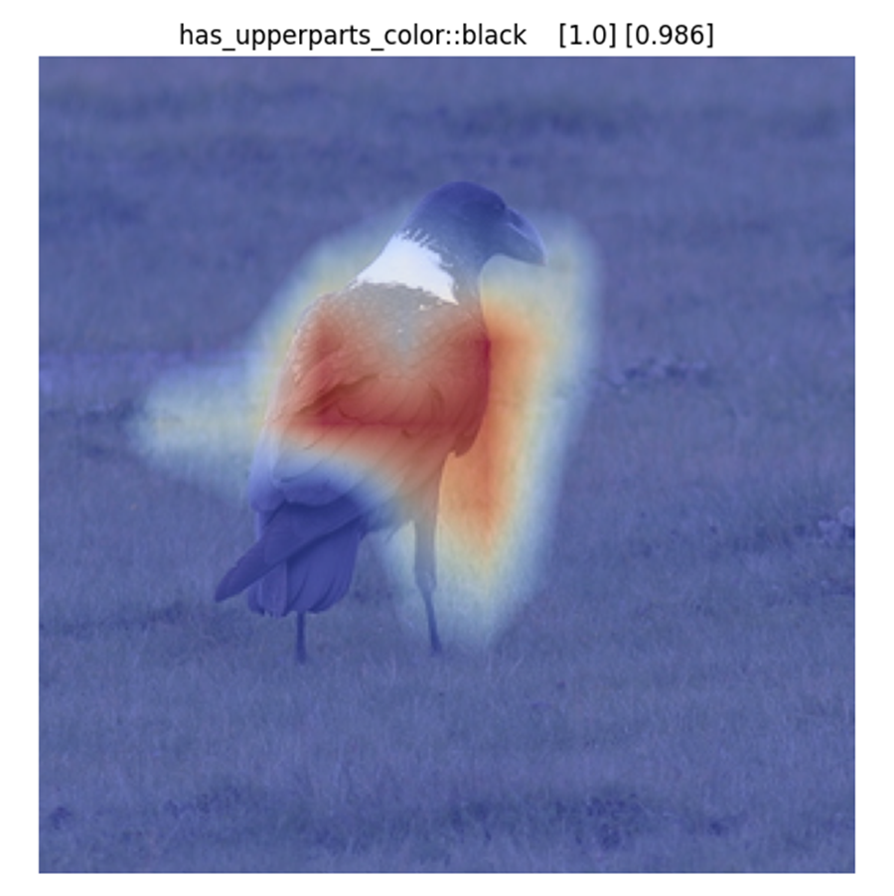} &
    \includegraphics[width=0.249\linewidth]{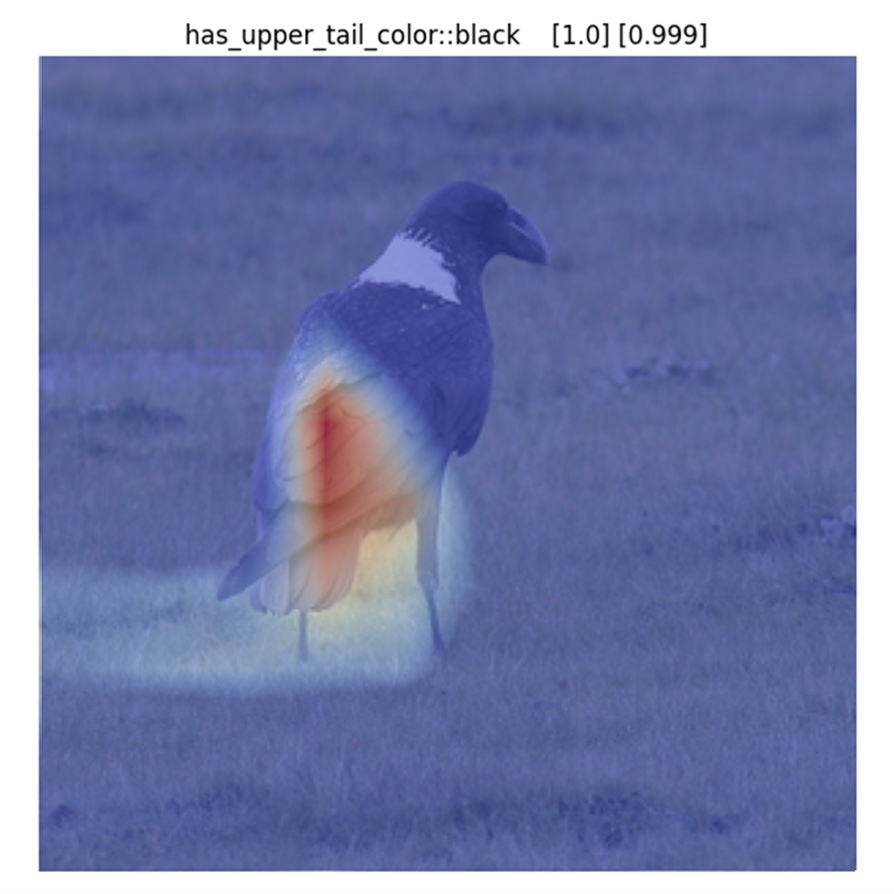} \\  
    \end{tabular*}}
    \vspace{-12pt}
    \caption{Concept saliency map on CUB dataset (white necked raven) shows reasonable localization of the ground truth concept regions in the input image. \label{fig:bird12}}
    \end{center}
\vspace{-12pt}
\end{figure*} 
\begin{figure*}[htbp] 
    \setlength{\tabcolsep}{0.2pt} % Default value: 6pt
    \renewcommand{\arraystretch}{1.0} % Default value: 1
    \begin{center}
    \resizebox{0.85\linewidth}{!}{
    \begin{tabular}
    {@{\extracolsep{\fill}}cccc}
    \begin{tabular}{c}
        \includegraphics[width=0.252\linewidth]{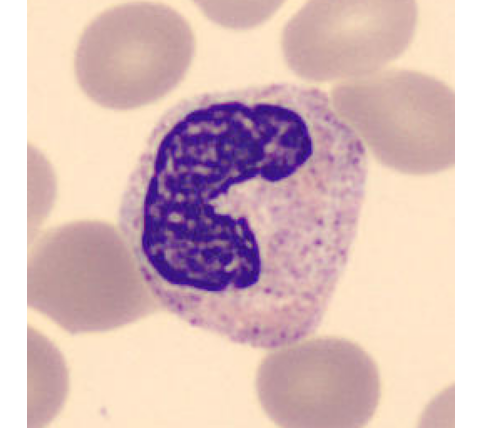} \\
        \footnotesize{(a) Original Image}
    \end{tabular} &
    \begin{tabular}{c}
        \includegraphics[width=0.247\linewidth]{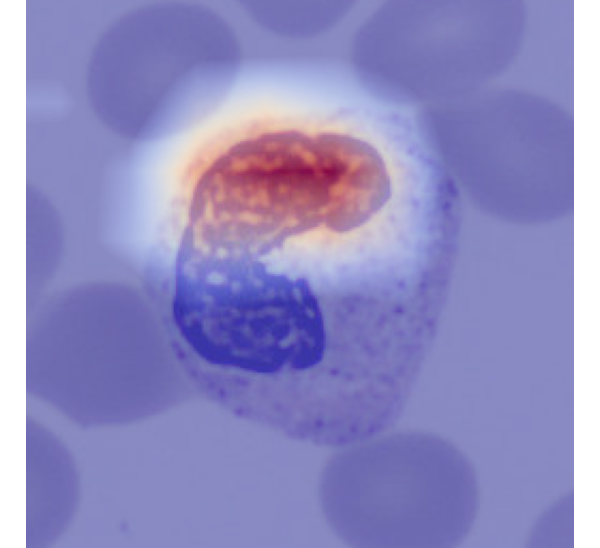} \\
        \footnotesize{(b) Granularity: Yes}
    \end{tabular} &
    \begin{tabular}{c}
        \includegraphics[width=0.247\linewidth]{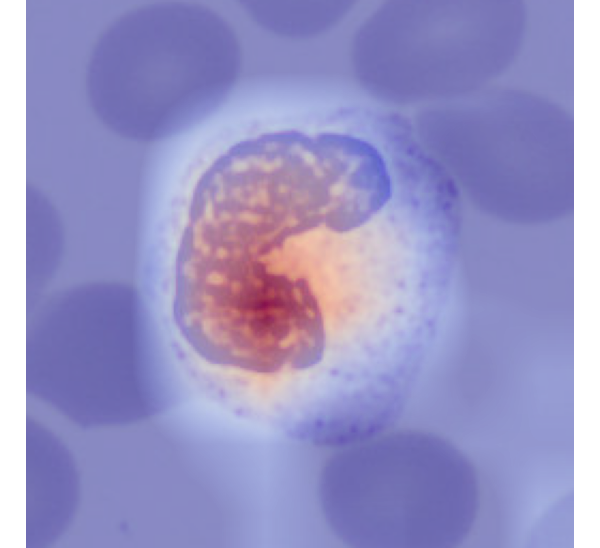} \\
        \footnotesize{(c) Cytoplasm Vacuole: No}
    \end{tabular} &
    \begin{tabular}{c}
        \includegraphics[width=0.247\linewidth]{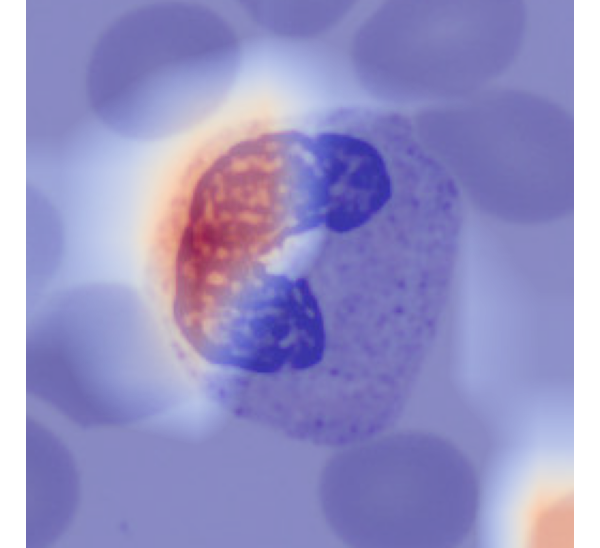} \\
        \footnotesize{(d) Chromatin Density: Density}
    \end{tabular} \\  
    \end{tabular}}

    \caption{
    Concept saliency map on WBCatt dataset (neutrophil) shows reasonable localization of the ground truth concept regions. \label{fig:cell2}}
    \end{center}

\end{figure*} 
\begin{figure*}[htbp] 
    \setlength{\tabcolsep}{0.2pt} % Default value: 6pt
    \renewcommand{\arraystretch}{1.0} % Default value: 1
    \begin{center}
    \resizebox{0.8\linewidth}{!}{
    \begin{tabular*}{\linewidth}
    {@{\extracolsep{\fill}}cccc}
    \includegraphics[width=0.249\linewidth]{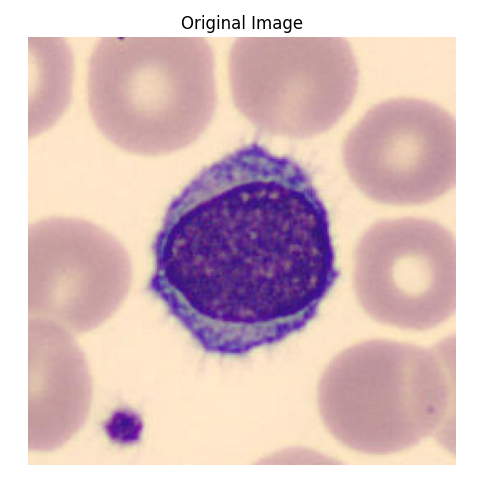} &
    \includegraphics[width=0.249\linewidth]{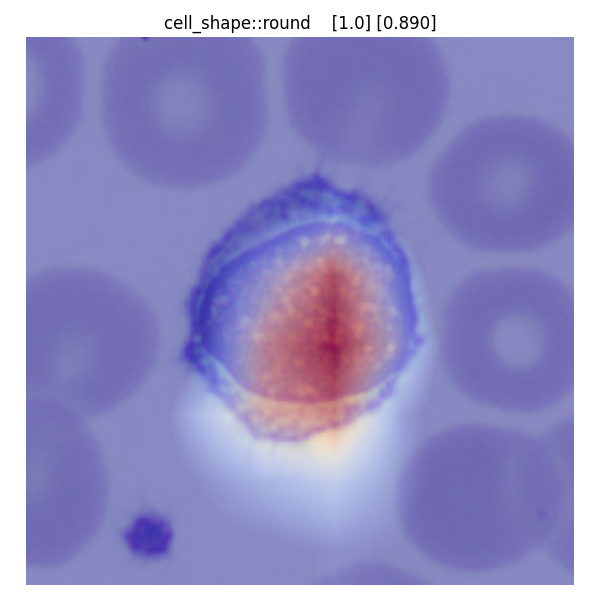} &
    \includegraphics[width=0.249\linewidth]{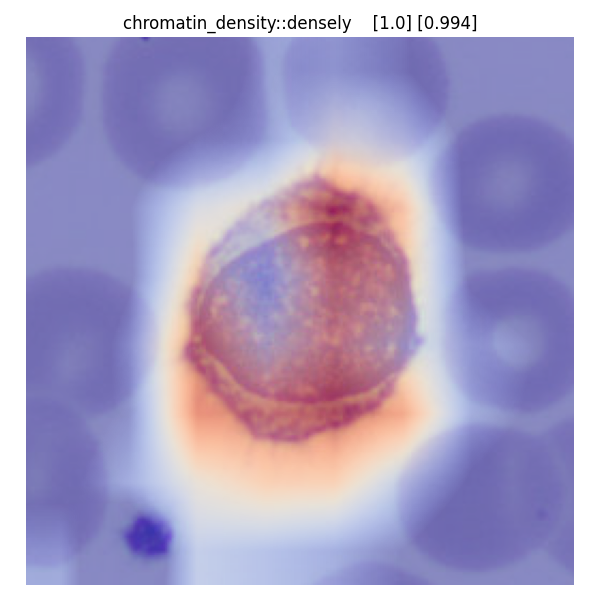} &
    \includegraphics[width=0.249\linewidth]{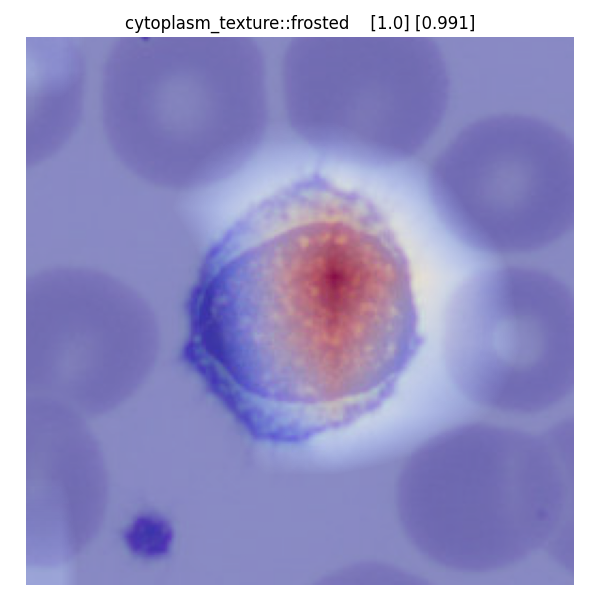} \\  
    \end{tabular*}}
    \vspace{-12pt}
    \caption{Concept saliency map on WBCatt dataset (lymphocyte) shows reasonable localization of the ground truth concept regions in the input image. \label{fig:cell4}}
    \end{center}
\vspace{-12pt}
\end{figure*} 
\begin{figure*}[htbp] 
    \setlength{\tabcolsep}{0.2pt} % Default value: 6pt
    \renewcommand{\arraystretch}{1.0} % Default value: 1
    \begin{center}
    \resizebox{0.8\linewidth}{!}{
    \begin{tabular*}{\linewidth}
    {@{\extracolsep{\fill}}cccc}
    \includegraphics[width=0.249\linewidth]{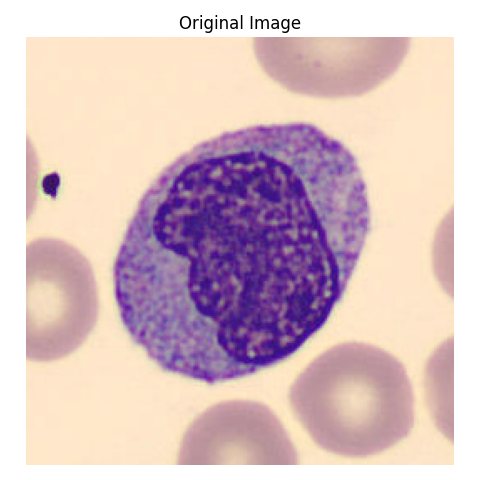} &
    \includegraphics[width=0.249\linewidth]{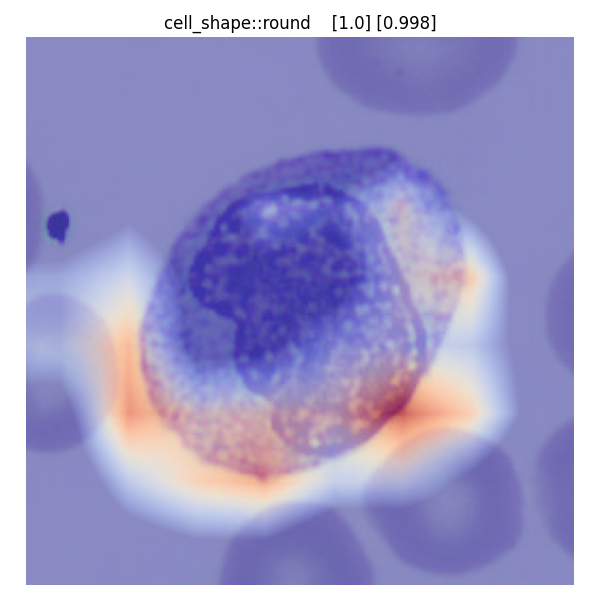} &
    \includegraphics[width=0.249\linewidth]{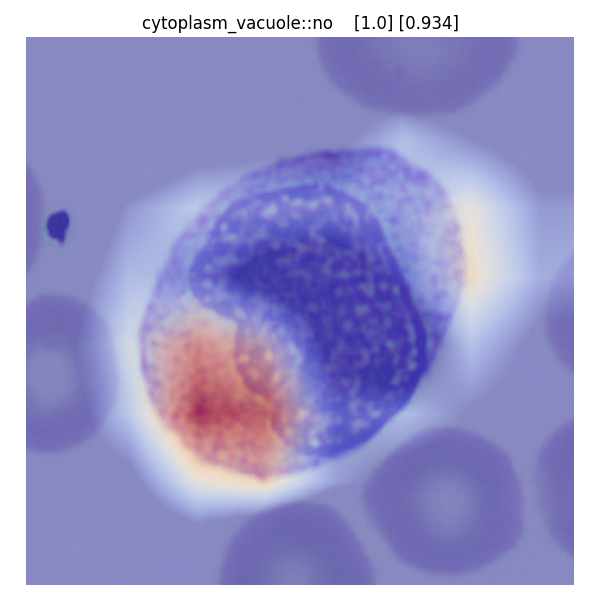} &
    \includegraphics[width=0.249\linewidth]{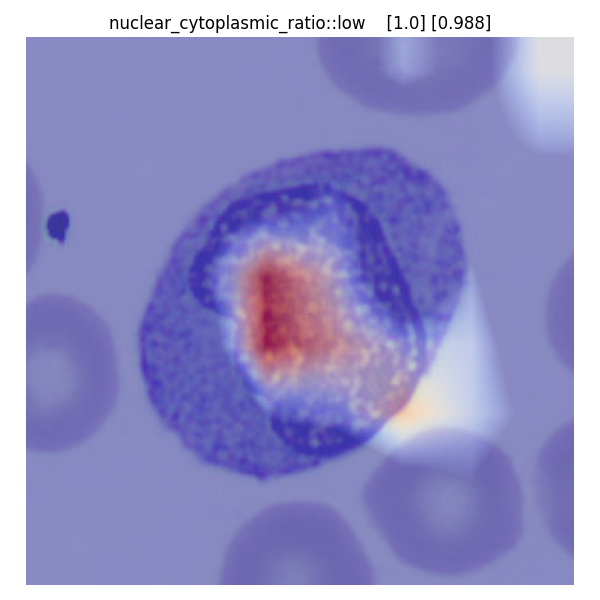} \\  
    \end{tabular*}}
    \vspace{-12pt}
    \caption{Concept saliency map on WBCatt dataset (monocyte) shows reasonable localization of the ground truth concept regions in the input image. \label{fig:cell5}}
    \end{center}
\vspace{-12pt}
\end{figure*} 
\begin{figure*}[htbp] 
    \setlength{\tabcolsep}{0.2pt} % Default value: 6pt
    \renewcommand{\arraystretch}{1.0} % Default value: 1
    \begin{center}
    \resizebox{0.8\linewidth}{!}{
    \begin{tabular*}{\linewidth}
    {@{\extracolsep{\fill}}cccc}
    \includegraphics[width=0.249\linewidth]{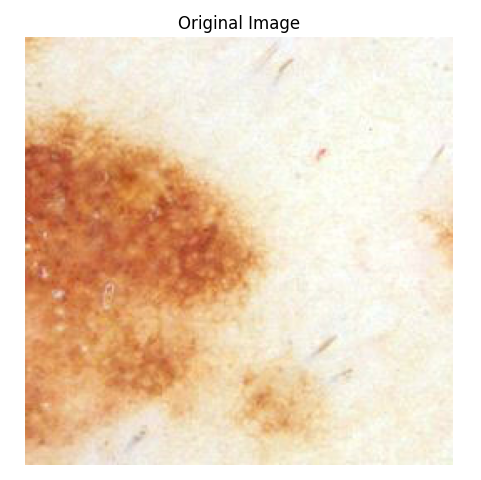} &
    \includegraphics[width=0.249\linewidth]{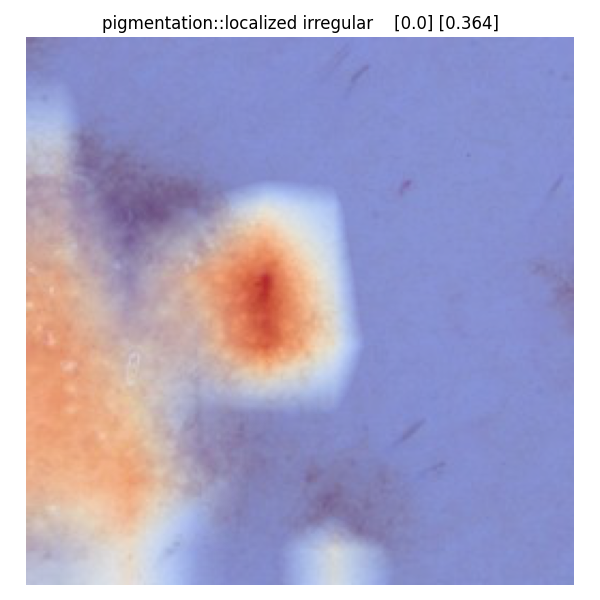} &
    \includegraphics[width=0.249\linewidth]{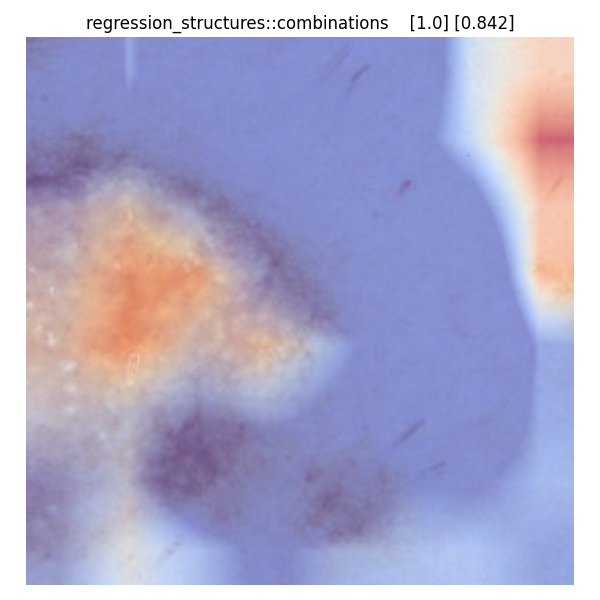} &
    \includegraphics[width=0.249\linewidth]{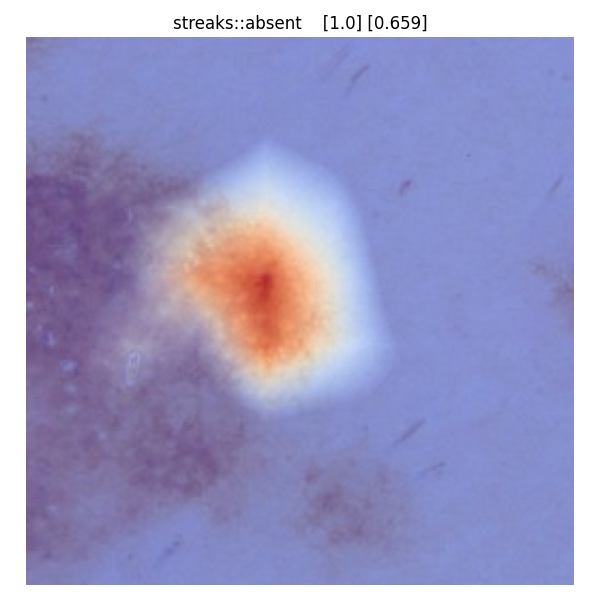} \\  
    \end{tabular*}}
    \vspace{-12pt}
    \caption{Concept saliency map on 7-point dataset (congenital nevus) shows reasonable localization of the ground truth concept regions in the input image. \label{fig:7point4}}
    \end{center}
\vspace{-12pt}
\end{figure*} 
\begin{figure*}[htbp] 
    \setlength{\tabcolsep}{0.2pt} % Default value: 6pt
    \renewcommand{\arraystretch}{1.0} % Default value: 1
    \begin{center}
    \resizebox{0.8\linewidth}{!}{
    \begin{tabular*}{\linewidth}
    {@{\extracolsep{\fill}}cccc}
    \includegraphics[width=0.249\linewidth]{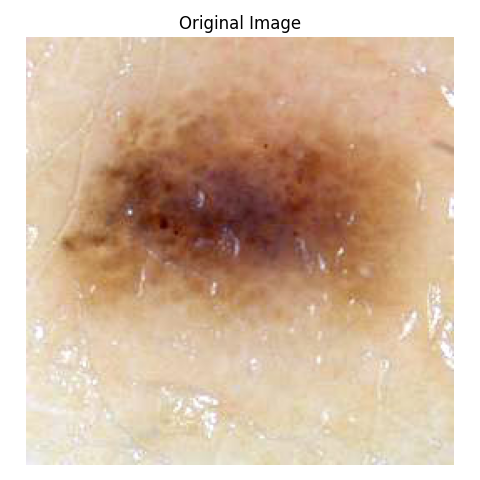} &
    \includegraphics[width=0.249\linewidth]{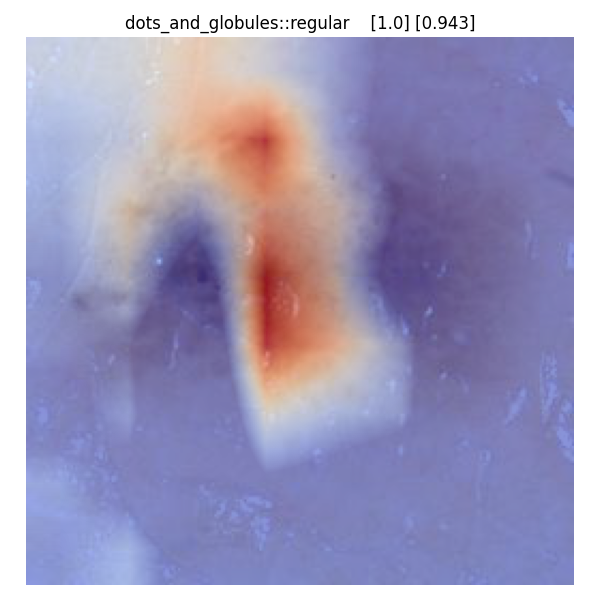} &
    \includegraphics[width=0.249\linewidth]{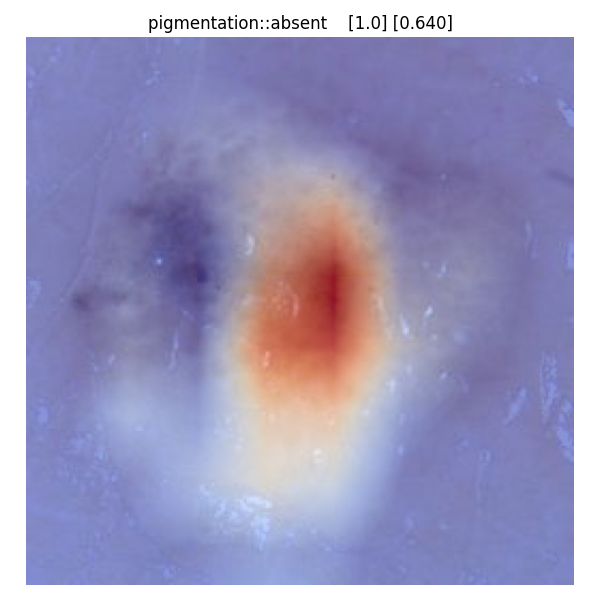} &
    \includegraphics[width=0.249\linewidth]{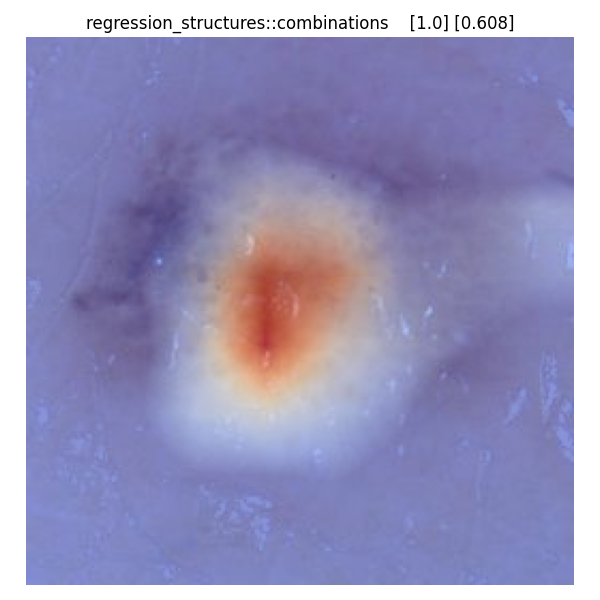} \\  
    \end{tabular*}}
    \vspace{-12pt}
    \caption{Concept saliency map on 7-point dataset (melanoma female) shows reasonable localization of the ground truth concept regions in the input image. \label{fig:7point5}}
    \end{center}
\vspace{-12pt}
\end{figure*} 

\section{Additional Interpretability Evaluation}\label{app:saliency_map}
We provide our additional interpretability evaluation in Figure \ref{fig:bird4} - \ref{fig:bird12} for CUB dataset, Figure \ref{fig:cell2} - \ref{fig:cell5} for WBCatt dataset, and Figure \ref{fig:7point4} - \ref{fig:7point5} for 7-point dataset as follows. Image regions that are highly relevant to the concept are highlighted.

\section{Limitations}
\label{sec:limitation}
While we solve a small portion of annotation problems by semi-supervised learning, semi-supervised models may not be suitable for all types of tasks or datasets. It is more effective that the data distribution is smooth. However, this is the limitation of semi-supervised learning, not our methods.

\section{Broader Impact}
\label{sec:impact}
The training of current CBMs heavily relies on the accuracy and richness of annotated concepts in the dataset. These concept labels are typically provided by experts, which can be costly and require significant resources and effort. Additionally, concept saliency maps frequently misalign with input saliency maps, causing concept predictions to correspond to irrelevant input features - an issue related to annotation alignment. In this problem, we propose SSCBM, a strategy to generate pseudo labels and an alignment loss to solve these two problems. Results show our effectiveness. This method has practical use in the real world.

\end{document}